\documentclass[11pt]{article}

\PassOptionsToPackage{table}{xcolor}
\usepackage{acl}
\usepackage{subfig}
\usepackage{times}
\usepackage{latexsym}
\usepackage{booktabs}
\usepackage{amsmath}
\usepackage{adjustbox}
\usepackage{times}
\usepackage{latexsym}
\usepackage{amsfonts}
\usepackage{float}
\usepackage{array}
\usepackage{siunitx}
\usepackage{multirow}
\usepackage[T1]{fontenc}

\usepackage[utf8]{inputenc}

\usepackage{microtype}

\usepackage{inconsolata}

\usepackage{graphicx}

\newcommand{\clipTwoTwoFour}{CLIP@224}
\newcommand{\clipThreeThreeSix}{CLIP@336}

\newcommand{\biovil}{BioViL-T}
\newcommand{\biomedclip}{BioMedCLIP}
\newcommand{\mrm}{MRM}
\newcommand{\raddinoControl}{RAD-DINO$_{\text{control}}$}
\newcommand{\raddino}{RAD-DINO}
\newcommand{\mimcxrvtwo}{MIMIC-CXR}
\newcommand{\radjepacontrol}{RadJEPA$_{\text{control}}$}
\newcommand{\radjepa}{RadJEPA}

\usepackage{graphicx,verbatim}

\title{RadJEPA: Radiology Encoder for Chest X-Rays via Joint
Embedding Predictive Architecture}

\author{
Anas Khan$^{1,4}$\thanks{Work done during Master's at IIT Bombay.} \quad
Mariam Husain$^{2}$\thanks{\textbf{Mariam Husain and Pratik Jalan contributed equally.}} \quad
Pratik Jalan$^{1,3}$\footnotemark[2] \quad
Kshitij Jadhav$^{3}$
\\
$^{1}$Department of Computer Science (CSE), Indian Institute of Technology Bombay, India \\
$^{2}$Department of Biomedical Engineering, Johns Hopkins University, USA \\
$^{3}$Koita Centre for Digital Health, Indian Institute of Technology Bombay, India 
\\
$^{4}$Mohamed bin Zayed University of Artificial Intelligence, Abu Dhabi, UAE
\\
\texttt{\{anaskhan, pratik\}@cse.iitb.ac.in \quad
anas.khan@mbzuai.ac.ae} \\
\texttt{mhusai10@jh.edu \quad
kshitij.jadhav@iitb.ac.in}
}

\begin{document}
\maketitle
\begin{abstract}

Vision-language pretraining has driven progress in medical image representation learning, but it depends on paired image-text data and can inherit reporting bias from clinical narratives. We study whether language-free predictive pretraining can produce an image encoder that transfers effectively to radiology report generation. RadJEPA is a chest-X-ray adaptation of I-JEPA, pretrained on approximately 840K unlabeled radiographs using latent 
context-to-target prediction. Our primary contribution is an extensive empirical evaluation of this language-free encoder for report generation: the frozen image encoder is coupled to a trainable two-layer projector and language decoder, and is also substituted into four established vision-language backbones. Across MIMIC-CXR and IU-Xray, RadJEPA matches or exceeds the evaluated image-only and image-text baselines on lexical, entity-relation, and 
clinical-label metrics. Controlled MIMIC-only comparisons provide evidence that the 
predictive objective contributes beyond domain-specific pretraining, while broader 
comparisons also reflect differences in pretraining data, model capacity, and input 
resolution. Complementary classification and segmentation experiments assess transfer 
beyond report generation. Code and pretrained weights are available at
\footnote{\href{https://huggingface.co/AIDElab-IITBombay/RadJEPA}
{RadJEPA Model Page}}.
\end{abstract}

\section{Introduction}
Vision--language modeling has largely been driven by the use of textual supervision \cite{radford2021learning}. In biomedical imaging \cite{zhou2023advancing}, representations are often reused as fixed visual tokens for downstream reasoning, such as report generation, image classification, and segmentation. This paradigm implicitly assumes that textual descriptions provide a sufficiently complete and unbiased account of visual content \cite{dehghani2023scaling}. More broadly, multimodal generation studies \cite{khan2024hope} have shown that  multimodal interpretation can be text-dominant. Radiology reports emphasize selective findings, frequently omitting subtle variations or absent observations. As a result, visual encoders trained to align images with text may inherit bias \cite{jones2024causal} and fail to preserve the full semantic.

Recent self-supervised alternatives such as DINO-style self-distillation have emerged as powerful label-free approaches ~\cite{caron2021emerging,oquab2023dinov2}. These methods learn invariances by aligning global representations across multiple augmented views of the same image using a teacher--student formulation in chest X-rays \cite{perez2025exploring}. Self-distillation methods \cite{dong2023maskclip} emphasize invariance by aligning representations across augmented views.


Chest X-rays are fundamentally \emph{semantic} rather than purely view-centric \cite{ccalli2021deep}: clinical interpretation depends on global anatomical context, spatial relationships, and subtle deviations from normal structure. Consequently, representation learning strategies that rely primarily on image--text alignment or view-level invariance may not explicitly model conditional dependencies within the image. Instead, we adopt a \emph{predictive latent modeling} perspective. 

Unlike contrastive alignment or self-distillation, this objective explicitly models conditional structure in representation space, encouraging abstraction of context-dependent semantic information without pixel reconstruction or textual supervision. JEPA formalize by predicting abstract latent representations of masked regions~\cite{assran2023self}. Latent predictive architectures have shown promise in neuroimaging~\cite{dong2024brain}, but they remain relatively underexplored in medical imaging.

\paragraph{Our main contributions are as follows:}




\begin{itemize}
    \item We adapt the I-JEPA latent predictive pretraining framework to chest X-rays and pretrain a ViT-B/14 encoder on 839,364 images from five open-source datasets, without report or label supervision.

    \item We provide an extensive empirical study of language-free visual pretraining for radiology report generation. We evaluate a frozen \radjepa{} encoder with a trainable projector and Vicuna-7B decoder and replace the native vision encoder in four established VLM pipelines, covering five language decoders in total.

    \item We compare against image-text, DINO-style, JEPA-style baselines; include data-matched control comparisons to separate objective effects from domain-pretraining effects; and report complementary classification and segmentation results under common downstream protocols.
\end{itemize}

\section{Related Works}
Chest X-ray representation learning methods can be broadly divided into image-text alignment and image-only self-supervised approaches. Vision-language models such as CLIP, BioViL, BiomedCLIP, CheXzero, and MRM learn joint image-text embeddings using paired radiographs and reports. CLIP \cite{radford2021learning} learns shared image-text embeddings using paired radiographs and reports. BioViL \cite{bannur2023learning} incorporates localized visual grounding, BiomedCLIP \cite{zhang2023biomedclip} scales biomedical contrastive pretraining, CheXzero \cite{tiu2022expert} enables zero-shot diagnosis generation, and MRM \cite{zhou2023advancing} improves multimodal representation matching. Medical image-conditioned language generation has also been explored beyond chest radiography. For example \cite{khan2024sumotosima}, combines visual representations with class-enriched textual information for otoscopic images.

Despite strong downstream performance, image-text contrastive frameworks have several limitations. They heavily rely on paired radiology reports, which often contain noise, reporting bias, and inconsistent terminology \cite{jones2024causal}. Furthermore, contrastive objectives mainly optimize global image-text alignment and may fail to capture fine-grained anatomical structure while also introducing shortcut correlations from language supervision \cite{sun2024exploring}.

To reduce dependence on language supervision, recent works explored image-only self-supervised learning methods. DINO-v2~\cite{oquab2023dinov2} learns representations through self-distillation between teacher and student networks using augmented views, while \raddino{}~\cite{perez2025exploring} adapts this framework specifically for chest X-ray representation learning using large-scale radiology datasets. Although these methods achieve strong downstream performance, they still rely heavily on augmentation consistency and explicit embedding alignment, which may encourage invariant shortcut features rather than contextual predictive understanding.

I-JEPA \cite{assran2023self} introduces predictive representation learning in latent space instead of reconstruction or contrastive alignment. Recent work such as CheXWorld \cite{yue2025chexworld} explored I-JEPA for chest X-ray representation learning; however, its evaluation was limited and did not include report generation or comprehensive AUPRC-based classification analysis. We compare it extensively in our work, with additional details provided in Section \ref{sec:CheXWorld}.

In contrast, our proposed RadJEPA adopts a JEPA-style predictive learning framework for radiology representation learning. Rather than aligning image-text embeddings or enforcing augmented-view consistency, RadJEPA learns contextual latent prediction directly from chest X-ray images without language supervision. This enables the model to focus more on structural and semantic understanding of radiological patterns while avoiding dependence on paired reports during pretraining.

\begin{table*}[t]
    \centering
    \footnotesize
    \begin{adjustbox}{width=\textwidth}
    \begin{tabular}{lcccccc}
        \toprule
        \textbf{Dataset} & \textbf{View} & \textbf{Patient cohort} &
        \textbf{\#Subjects} &
        \textbf{\#Images} &
        \textbf{Frontal} &
        \textbf{Lateral} \\
        \midrule
        BRAX~\cite{reis2022brax}
            & frontal, lateral & institutional PACS 
            & 19,351 & 41,620 & 24,959 & 16,661 \\

        CheXpert~\cite{irvin2019chexpert} 
            & frontal, lateral & inpatient, outpatient 
            & 65,240 & 224,316 & 191,229 & 33,087 \\

        MIMIC-CXR~\cite{johnson2019mimic} 
            & frontal, lateral & ICU 
            & 188,546 & 300,491 & 210,491 & 90,000 \\

        ChestX-ray14~\cite{wang2017chestx} 
            & frontal & not specified 
            & 32,717 & 112,120 & 112,120 & 0 \\

        PadChest~\cite{bustos2020padchest} 
            & frontal, lateral & all available 
            & 67,000 & 160,817 & 96,287 & 64,530 \\

        \midrule
        \textbf{Total} & & &
        \textbf{372,854} &
        \textbf{839,364} &
        \textbf{635,086} &
        \textbf{204,278} \\
        \bottomrule
    \end{tabular}
    \end{adjustbox}
    \caption{Chest X-ray datasets used for RadJEPA pretraining. For MIMIC-CXR, only a subset of subjects is included to avoid overlap with the evaluation sets.}
    \label{tab:pretrain_data}
\end{table*}

\begin{table*}[t]
    \centering
    \footnotesize
    \resizebox{\textwidth}{!}{%
    \begin{tabular}{@{}lllrlrrc@{}}
        \toprule
        \textbf{Model type} & \textbf{Model} & \textbf{Arch.} & \textbf{\# Params.} &
        \textbf{Training dataset} & \textbf{\# Images} & \textbf{\# Text} & \textbf{Resolution} \\
        \midrule
        Image \& Text & \clipTwoTwoFour~\cite{radford2021learning}     & ViT-L/14 & 304 M & WebImageText & 400 M & 400 M & 224$^2$ \\
        Image \& Text & \clipThreeThreeSix~\cite{radford2021learning} & ViT-L/14 & 304 M & WebImageText & 400 M & 400 M & 336$^2$ \\
        Image \& Text & \biovil~\cite{bannur2023learning}              & ResNet50 & 27 M  & \mimcxrvtwo  & 197 k & 174 k & 512$^2$ \\
        Image \& Text & \biomedclip~\cite{zhang2023biomedclip}              & ViT-B/16 & 86 M  & PMC-15M      & 15 M  & 15 M  & 224$^2$ \\
        Image \& Text & CheXzero~\cite{tiu2022expert}                  & ViT-B/32 & 151 M & \mimcxrvtwo  & 377 k & 227 k & 224$^2$ \\
        Image \& Text & \mrm~\cite{zhou2023advancing}                  & ViT-B/16 & 86 M  & \mimcxrvtwo  & 377 k & 227 k & 448$^2$ \\
        \midrule
        Image Only & CheXWorld~\cite{yue2025chexworld}                   & ViT-B/14 & 86M & multiple datasets          & 448K & --    & 224$^2$ \\
        Image Only & DINO-v2~\cite{oquab2023dinov2}                   & ViT-G/14 & 1.1 B & LVD          & 142 M & --    & 518$^2$ \\
        Image Only & DINO-v3~\cite{simeoni2025dinov3} & ViT-B/16 & 86 M & LVD-1689M & 1.689 B & -- & 224$^2$ \\
        Image Only & \raddinoControl                                  & ViT-B/14 & 87 M  & \mimcxrvtwo  & 197 k & --    & 518$^2$ \\
        Image Only & \raddino~\cite{perez2025exploring}                     & ViT-B/14 & 87 M  & Multi-CXR    & 838 k & --    & 518$^2$ \\
        \midrule
        Image Only & I-JEPA~\cite{assran2023self}              & ViT-H/14 & 0.6 B & IN1K         & 1.28 M & --    & 448$^2$ \\
        Image Only & \radjepacontrol                            & ViT-B/14 & 86 M  & \mimcxrvtwo  & 197 k & --    & 224$^2$ \\
        Image Only & \radjepa                                   & ViT-B/14 & 86 M  & Multi-CXR    & 839 k & --    & 224$^2$ \\
        \bottomrule
    \end{tabular}%
    }
    \caption{Overview of image backbones and their training dataset characteristics employed in experimental analysis.}
    \label{tab:all_model_details}
\end{table*}

\section{Pretraining Setup and Baselines}

\begin{figure*}[t]
    \centering
    \includegraphics[width=\linewidth]{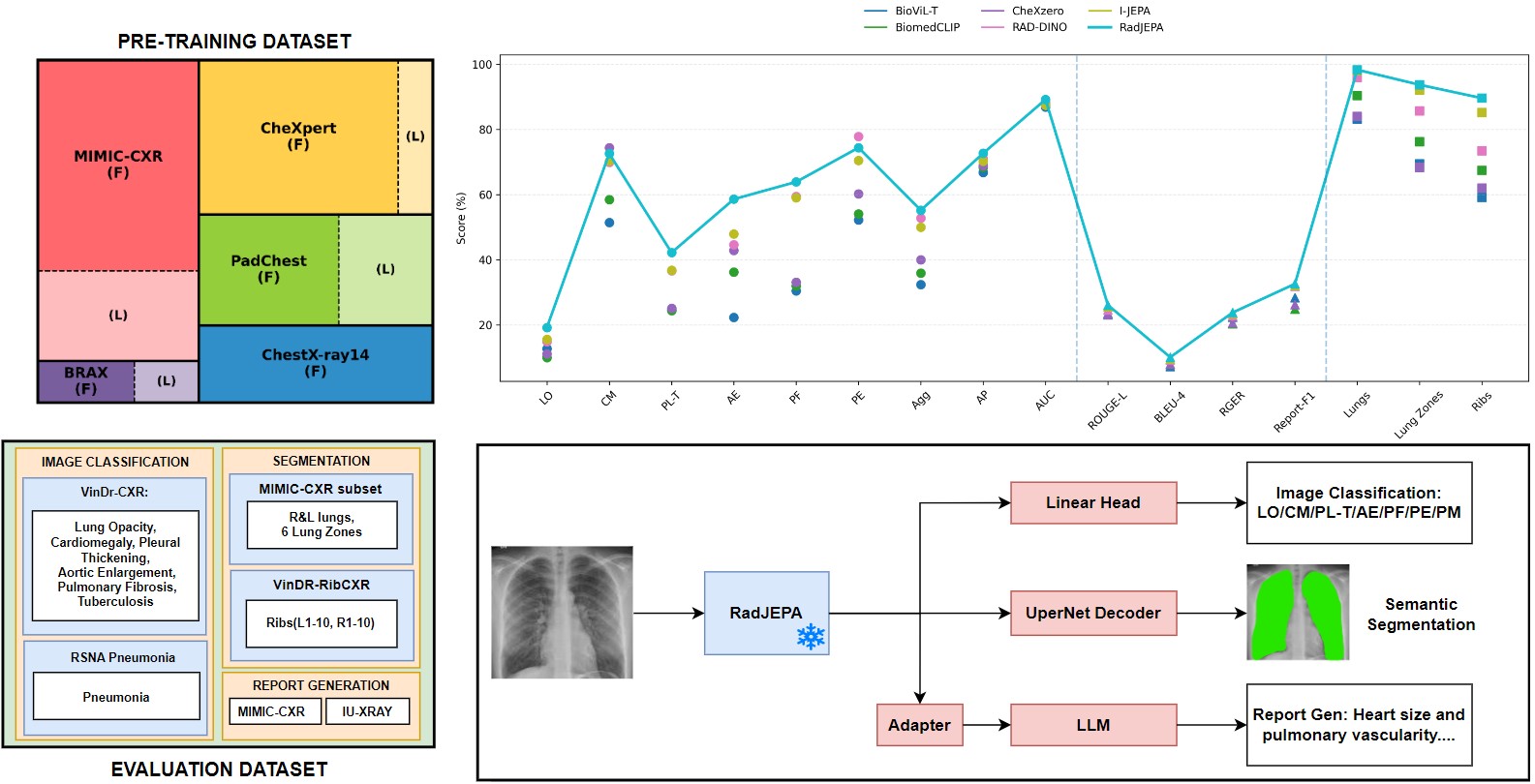}
    \caption{
Overview of the RadJEPA framework and evaluation.
\textbf{Top-left:} Proportional visualization of pre-training datasets, where area reflects dataset size; (F) and (L) denote frontal and lateral views, respectively.
\textbf{Bottom-left:} Summary of evaluation datasets across image classification, segmentation, and report generation.
\textbf{Bottom-right:} Model pipeline during downstream evaluation, RadJEPA is frozen, while task-specific heads (linear head, UPerNet decoder, adapter + LLM) are trained.
\textbf{Top-right:} Unified performance comparison across three downstream tasks, reporting five baselines consistently evaluated across all tasks.
}
    \label{fig:model_overview}
\end{figure*}

\radjepa{} pretraining follows the I-JEPA protocol~\cite{assran2023self}, as described in Section~\ref{sec:radjepa_methodology} and Figure~\ref{fig:model_overview} (Parameters details c.f. Table \ref{tab:pretraining_optimization_comparison}). We adopt a data configuration aligned with \raddino{}~\cite{perez2025exploring}, leveraging large-scale chest X-ray datasets summarised in Table~\ref{tab:pretrain_data}. In publicly available datasets, frontal views substantially outnumber lateral views (approximately $6{:}1$). Unlike \raddino{}, which incorporates approximately $90$K proprietary images, we rely exclusively on open-source data. To mitigate viewpoint imbalance, we augment the lateral subset with approximately $90$K additional lateral chest X-rays from MIMIC-CXR, giving a final frontal-to-lateral ratio of approximately $3{:}1$. In total, \radjepa{} is pretrained on $839{,}364$ chest X-ray images. A set of baseline approaches is considered for experimental analysis (Table~\ref{tab:all_model_details}), with \raddino{} serving as the key image-only baseline. Comparison with other generic image-only self-supervised methods is outside the scope of this study, as it has been extensively explored in prior work~\cite{oquab2023dinov2,caron2021emerging}. As summarised in Table~\ref{tab:all_model_details}, \radjepa{} uses a ViT-B/14 backbone; despite its smaller capacity, this encoder consistently outperforms substantially larger image-only backbones in the downstream evaluations.

\section{Methodology}
\label{sec:radjepa_methodology}


Let $x \in \mathbb{R}^{H \times W}$ denote a chest X-ray. RadJEPA pretraining operates by sampling a context region $c$ and multiple target regions $t$. Visual encoder $f$ extracts latent representations $z_c = f(c)$ and $z_t = f(t)$, a predictor $g$ estimates the target embedding as $\hat{z}_t = g(z_c)$. Model is trained using a latent prediction objective
$\mathcal{L}_{\text{JEPA}} = \mathbb{E}_{(c,t)} \left[ \operatorname{SmoothL1}\left(\text{stopgrad}(z_t), \hat{z}_t\right) \right]$.
The target encoder parameters $\theta'$ are updated via EMA $\theta' \leftarrow \tau \theta' + (1-\tau)\theta$, where $\tau \in [0,1)$. After pretraining, the encoder $f$ is frozen and reused across downstream tasks (cf.\ Figure \ref{fig:model_overview}).

\textbf{Disease Classification.}
Given an image $x$, frozen features $h = f(x)$ are passed to a linear classifier producing probabilities $p = \text{softmax}(Wh)$, optimized using cross-entropy 
$\mathcal{L}_{\text{cls}} = - \sum_{k=1}^{K} y_k \log p_k$.
Only the classification head is trained.

\textbf{Semantic Segmentation.}
Multi-scale features $\{h_l\}_{l=1}^{L} = f^{(l)}(x)$ are extracted and aggregated by a UperNet decoder $\mathcal{D}$ to produce pixel-wise logits $S = \mathcal{D}(\{h_l\})$. The segmentation loss is 
$\mathcal{L}_{\text{seg}} = - \sum_{i,j} \sum_{c=1}^{C} y_{ij}^c \log \text{softmax}(S_{ij}^c)$.
Only decoder parameters are optimized.

\textbf{Report Generation.}
Frozen visual embeddings $v = f(x)$ are projected into the language space through a two-layer MLP adapter $\tilde{v} = A(v) = v + \lambda\, W_2\, \sigma(W_1 v)$, where $\sigma$ is the GeLU non-linearity and $\lambda$ is a learnable residual scalar initialised to $1$. The projected embeddings are concatenated with an instruction prompt and passed to a language model $\mathcal{M}$ (Vicuna-7B in the main experiments), which generates reports autoregressively with $p(y_t \mid y_{<t}, x) = \mathcal{M}(y_t \mid y_{<t}, \tilde{v})$ and is optimised via negative log-likelihood $\mathcal{L}_{\text{txt}} = - \sum_{t=1}^{T} \log p(y_t \mid y_{<t}, x)$. The encoder $f$ remains frozen; only $A$ and $\mathcal{M}$ are updated.

\begin{table*}[t]
    \centering
    \small
    \begin{adjustbox}{width=\textwidth}
    \begin{tabular}{
        llc
        >{\centering\arraybackslash}S[table-format=4]
        *{4}{>{\centering\arraybackslash}p{2.1cm}}
    }
        \toprule
        \textbf{Dataset} & \textbf{Image encoder} & \textbf{Input res.} & {\# Tokens}
        & ROUGE-L & BLEU-4 & RG\textsubscript{ER} & Macro-F1-14 \\
        \midrule

        \multirow{10}{*}{\textbf{MIMIC}}
       & \clipTwoTwoFour \cite{radford2021learning}
            & 224$\times$224 & 256
            & 23.0 [22.7, 23.4]
            & 8.3 [7.9, 8.6]
            & 20.3 [19.8, 20.7]
            & 24.7 [23.6, 26.0] \\
        
        & \clipThreeThreeSix \cite{radford2021learning}
            & 336$\times$336 & 576
            & 23.3 [22.9, 23.7]
            & 8.4 [8.0, 8.7]
            & 20.4 [19.9, 20.9]
            & 25.3 [24.2, 26.5] \\
        
        & DINO-v2 \cite{oquab2023dinov2}
            & 518$\times$518 & 261
            & 22.7 [22.4, 23.2]
            & 7.6 [7.3, 7.9]
            & 18.5 [18.1, 19.1]
            & 18.6 [17.8, 19.5] \\
        & DINOv3 \cite{simeoni2025dinov3}
            & 224$\times$224 & 201
            & 21.4 [21.0, 21.8]
            & 7.3 [7.0, 7.7]
            & 18.8 [18.3, 19.2]
            & 18.1 [17.2, 19.0] \\
        
        \cmidrule(lr){2-8}
        
        & BiomedCLIP \cite{zhang2023biomedclip}
            & 224$\times$224 & 256
            & 23.1 [22.8, 23.5]
            & 7.9 [7.5, 8.2]
            & 20.4 [19.9, 20.8]
            & 24.9 [23.8, 26.1] \\
        
        & CheXzero \cite{tiu2022expert}
            & 224$\times$224 & 49
            & 23.2 [22.9, 23.6]
            & 8.0 [7.7, 8.4]
            & 20.6 [20.2, 21.1]
            & 26.2 [25.0, 27.5] \\
        
        & \biovil \cite{bannur2023learning}
            & 512$\times$512 & 196
            & 23.5 [23.2, 23.9]
            & 7.3 [7.0, 7.6]
            & 22.4 [21.9, 22.8]
            & 28.4 [27.2, 29.8] \\
        
        & \raddinoControl
            & 518$\times$518 & 1369
            & 24.2 [23.8, 24.6]
            & 9.0 [8.7, 9.4]
            & 22.4 [21.9, 22.9]
            & 31.5 [30.1, 32.9] \\
        
        & \raddino~\cite{perez2025exploring}
            & 518$\times$518 & 1369
            & 25.1 [24.7, 25.4]
            & 9.4 [9.0, 9.8]
            & 23.2 [22.7, 23.6]
            & 32.0 [30.5, 33.5] \\
        \cmidrule(lr){2-8}
        & CheXWorld~\cite{yue2025chexworld}
            & 224$\times$224 & 256
            & 24.1 [23.6, 24.5]
            & 8.9 [8.4, 9.3]
            & 20.1 [19.7, 20.4]
            & 30.2 [28.8, 31.7] \\
        & I-JEPA~\cite{assran2023self}
            & 224$\times$224 & 256
            & 23.2 [22.8, 23.6]
            & 8.1 [7.7, 8.5]
            & 21.6 [21.2, 22.0]
            & 29.4 [27.9, 31.0] \\
        
        & RadJEPA\textsubscript{control}
            & 224$\times$224 & 256
            & 25.5 [25.1, 25.8]
            & 9.7 [9.3, 10.0]
            & 23.3 [22.9, 23.7]
            & 31.9 [30.7, 33.2] \\
        \rowcolor{gray!15}
        & RadJEPA
            & 224$\times$224 & 256
            & \textbf{26.1 [25.7, 26.4]}
            & \textbf{10.1 [9.7, 10.4]}
            & \textbf{23.8 [23.4, 24.2]}
            & \textbf{32.6 [31.2, 33.8]} \\

        \midrule

        \multirow{10}{*}{\textbf{IU}}
        & \clipTwoTwoFour \cite{radford2021learning}
            & 224$\times$224 & 256
            & 25.4 [25.1, 25.7]
            & 9.2 [8.9, 9.5]
            & 25.8 [25.3, 26.2]
            & 18.1 [16.1, 20.8] \\
        
        & \clipThreeThreeSix \cite{radford2021learning}
            & 336$\times$336 & 576
            & 25.3 [24.9, 25.6]
            & 8.0 [7.8, 8.3]
            & 25.3 [24.8, 25.6]
            & 18.5 [16.7, 20.8] \\
        
        & DINO-v2 \cite{oquab2023dinov2}
            & 518$\times$518 & 261
            & 25.4 [25.1, 25.7]
            & 8.0 [7.7, 8.2]
            & 23.6 [23.2, 24.0]
            & 12.3 [10.6, 14.1] \\

        & DINOv3 \cite{simeoni2025dinov3}
            & 224$\times$224 & 201
            & 25.3 [24.9, 25.7]
            & 7.8 [7.5, 8.1]
            & 23.9 [23.6, 24.0]
            & 12.5 [11.7, 13.4] \\
        
        \cmidrule(lr){2-8}
        
        & BiomedCLIP \cite{zhang2023biomedclip}
            & 224$\times$224 & 256
            & 20.2 [19.9, 20.4]
            & 6.3 [6.1, 6.5]
            & 20.0 [19.7, 20.4]
            & 7.1 [5.9, 8.5] \\
        
        & CheXzero \cite{tiu2022expert}
            & 224$\times$224 & 49
            & 25.6 [25.2, 25.9]
            & 8.5 [8.2, 8.8]
            & 25.7 [25.2, 26.1]
            & 18.1 [16.3, 20.1] \\
        
        & \biovil \cite{bannur2023learning}
            & 512$\times$512 & 196
            & 26.3 [25.9, 26.6]
            & 8.2 [7.9, 8.4]
            & 25.3 [24.9, 25.7]
            & 20.2 [18.0, 23.0] \\
        
        & \raddinoControl
            & 518$\times$518 & 1369
            & 25.5 [25.2, 25.9]
            & 9.2 [8.9, 9.4]
            & 26.2 [25.8, 26.6]
            & 23.8 [21.4, 26.3] \\
        
        & \raddino~\cite{perez2025exploring}
            & 518$\times$518 & 1369
            & 26.3 [25.9, 26.7]
            & 9.4 [9.1, 9.8]
            & 26.6 [26.1, 27.0]
            & 26.4 [24.0, 28.7] \\
        \cmidrule(lr){2-8}
        & CheXWorld~\cite{yue2025chexworld}
            & 224$\times$224 & 256
            & 26.1 [25.8, 26.5]
            & 9.1 [8.6, 9.6]
            & 26.3 [25.8, 26.9]
            & 24.9 [23.4, 26.5] \\
        & I-JEPA~\cite{assran2023self}
            & 224$\times$224 & 256
            & 25.8 [25.3, 26.2]
            & 8.9 [8.5, 9.2]
            & 25.9 [25.4, 26.4]
            & 25.4 [23.1, 27.8] \\
        
        & RadJEPA\textsubscript{control}
            & 224$\times$224 & 256
            & 27.1 [26.7, 27.4]
            & 9.6 [9.3, 9.9]
            & 27.0 [26.5, 27.4]
            & 26.8 [24.8, 29.1] \\
        \rowcolor{gray!15}
        & RadJEPA
            & 224$\times$224 & 256
            & \textbf{28.4 [28.0, 28.7]}
            & \textbf{9.9 [9.6, 10.2]}
            & \textbf{27.5 [27.1, 27.9]}
            & \textbf{27.6 [25.3, 29.8]} \\

        \bottomrule
    \end{tabular}
    \end{adjustbox}
    \caption{
    Downstream radiology report generation results on MIMIC-CXR and IU-Xray. In all models, the image encoder remains frozen, while the two-layer MLP projector and the language decoder are fully fine-tuned. We report median and 95\% confidence intervals from $500$ bootstrap samples. \radjepacontrol{} denotes the variant pretrained only on MIMIC-CXR for an architecture-matched comparison with \raddinoControl{} (cf. Table~\ref{tab:all_model_details}). Numbers in bold are the best in each column. The statistical protocol is described in Section~\ref{sec:significance}; complementary visual analysis is in Section~\ref{sec:appendix_analysis}.
    }
    \label{tab:findings_generation}
\end{table*}

\section{Evaluation}
\paragraph{Classification and Segmentation: } 
We also evaluated RadJEPA on classification and segmentation tasks. Although these tasks are not directly related to computational linguistics, as they do not involve textual modalities, they are important for understanding the potential and generalization capability of RadJEPA on foundational vision tasks. Note that we could not evaluate performance on the CANDID-PTX dataset \cite{feng2021curation}, as, upon contacting the authors, the dataset is no longer publicly available as it was previously. A detailed comparison is provided in Section \ref{sec:class_seg}.

\subsection{Compare with Existing Vision Encoders}

\subsubsection*{Datasets and Implementation Details}
\label{sec:report_gen}
We evaluate report generation on \textbf{MIMIC-CXR} and \textbf{IU-Xray}. For MIMIC-CXR we follow the official splits used in \raddino{}~\cite{perez2025exploring}, retaining only frontal chest X-rays with a \textit{Findings} section, yielding $146{,}909$/$7{,}250$/$2{,}461$ train/validation/test image-text pairs. These splits are used to fine-tune the language decoder; the image encoder remains frozen. To probe out-of-distribution generalisation we additionally report results on IU-Xray ($3{,}306$ test studies), which is not used during the training of either the encoder or the decoder. The pipeline follows the LLaVA-1.5 recipe~\cite{liu2024improved}: patch embeddings from the frozen image encoder are projected through a two-layer MLP and concatenated with the instruction prompt ``$\langle$image\_tokens$\rangle$ Provide a description of the findings in the radiology image.''. We use Vicuna-7B (v1.5) as the language model $\mathcal{M}$. Training uses AdamW with cosine scheduling at peak learning rate $2{\times}10^{-5}$, batch size $128$ ($32$ per GPU) with $3\%$ warmup, on four NVIDIA A100 nodes for three epochs. During inference, we use 32-bit precision, up to $150$ tokens, and a batch size of $1$.

We report standard report-generation metrics. \textbf{ROUGE-L}~\cite{lin2004rouge} and \textbf{BLEU-4}~\cite{papineni2002bleu} measure $n$-gram overlap with the reference \textit{Findings}. \textbf{RG\textsubscript{ER}} denotes the entity-relation F1 score from the RadGraph parser of~\citet{yu2023evaluating, jain2021radgraphextractingclinicalentities} and captures the alignment of clinical entities and their relations. \textbf{Macro-F1-14} is the macro-averaged F1 of the $14$ CheXpert-style labels extracted from the generated and reference reports by the CheXbert labeller~\cite{smit2020chexbert}.

\begin{table*}[!ht]
    \centering
    \small
    \begin{adjustbox}{width=\textwidth}
    \begin{tabular}{
        >{\centering\arraybackslash}m{2.2cm}|
        >{\centering\arraybackslash}m{2.2cm}
        >{\centering\arraybackslash}m{2.2cm}
        *{4}{>{\centering\arraybackslash}m{2.2cm}}
    }
        \toprule
        \textbf{Dataset} & \textbf{VLMs} & \textbf{Encoder}
        & \textbf{ROUGE-L} & \textbf{BLEU-4} & \textbf{RG\textsubscript{ER}} & \textbf{Macro-F1-14} \\
        \midrule

        \multirow{20}{*}{\textbf{MIMIC-CXR}}

        & \multirow{5}{*}{MedLLaVA}
            & Base Encoder
            & 21.4 [21.0, 21.8]
            & 5.3 [5.0, 5.6]
            & 16.8 [16.4, 17.2]
            & 18.9 [17.1, 20.7] \\
        &   & I-JEPA
            & 24.9 [24.4, 25.4]
            & 9.1 [8.7, 9.5]
            & 19.9 [19.4, 20.4]
            & 27.4 [25.5, 29.5] \\
        &   & DINO-v2
            & 22.9 [22.5, 23.3]
            & 7.8 [7.5, 8.1]
            & 18.3 [17.9, 18.8]
            & 14.7 [13.0, 16.5] \\
        &   & DINO-v3
            & 22.3 [21.9, 22.7]
            & 7.6 [7.3, 8.0]
            & 18.0 [17.6, 18.4]
            & 14.6 [12.9, 16.4] \\
        &   & RAD-DINO
            & 25.2 [24.8, 25.6]
            & 9.0 [8.7, 9.4]
            & 21.9 [21.4, 22.3]
            & 29.2 [27.7, 30.9] \\
        &   & RadJEPA
            & \textbf{25.9 [25.6, 26.2]}
            & \textbf{9.6 [9.3, 9.9]}
            & \textbf{22.4 [22.1, 22.8]}
            & \textbf{29.8 [28.5, 31.1]} \\
        
        \cmidrule(lr){2-7}
        
        & \multirow{5}{*}{Qwen-2.5}
            & Base Encoder
            & 22.9 [22.5, 23.3]
            & 5.1 [4.8, 5.4]
            & 18.5 [18.0, 18.9]
            & 16.6 [14.8, 18.5] \\
        &   & I-JEPA
            & 27.2 [26.7, 27.7]
            & 10.2 [9.8, 10.6]
            & 22.5 [22.0, 23.0]
            & 29.8 [27.9, 31.8] \\
        &   & DINO-v2
            & 24.0 [23.6, 24.4]
            & 8.4 [8.1, 8.7]
            & 18.7 [18.2, 19.2]
            & 16.8 [14.9, 18.7] \\
        &   & DINO-v3
            & 23.7 [23.3, 24.1]
            & 8.5 [8.1, 8.8]
            & 18.8 [18.3, 19.3]
            & 16.2 [14.4, 18.1] \\
        &   & RAD-DINO
            & 26.8 [26.4, 27.2]
            & 9.9 [9.5, 10.3]
            & 22.8 [22.3, 23.3]
            & 31.3 [29.6, 33.1] \\
        &   & RadJEPA
            & \textbf{28.1 [27.8, 28.4]}
            & \textbf{10.7 [10.4, 11.0]}
            & \textbf{23.5 [23.2, 23.9]}
            & \textbf{32.9 [31.4, 34.2]} \\
        
        \cmidrule(lr){2-7}
        
        & \multirow{5}{*}{BLIP-2}
            & Base Encoder
            & 23.0 [22.6, 23.4]
            & 5.6 [5.3, 5.9]
            & 17.8 [17.4, 18.2]
            & 18.5 [16.6, 20.4] \\
        &   & I-JEPA
            & 25.7 [25.2, 26.2]
            & 9.1 [8.7, 9.5]
            & 23.9 [23.4, 24.4]
            & 24.7 [22.9, 26.7] \\
        &   & DINO-v2
            & 24.3 [23.9, 24.7]
            & 7.9 [7.6, 8.2]
            & 19.8 [19.3, 20.3]
            & 15.3 [13.5, 17.1] \\
        &   & DINO-v3
            & 24.0 [23.6, 24.4]
            & 7.6 [7.3, 7.8]
            & 19.7 [19.2, 20.3]
            & 14.9 [13.1, 16.8] \\
        &   & RAD-DINO
            & 26.5 [26.1, 26.9]
            & 9.4 [9.0, 9.8]
            & 23.5 [23.0, 24.0]
            & 29.9 [28.2, 31.6] \\
        &   & RadJEPA
            & \textbf{27.9 [27.6, 28.2]}
            & \textbf{10.3 [10.0, 10.6]}
            & \textbf{24.1 [23.7, 24.4]}
            & \textbf{31.7 [30.2, 33.0]} \\
        
        \cmidrule(lr){2-7}
        
        & \multirow{5}{*}{Phi-4}
            & Base Encoder
            & 18.2 [17.8, 18.6]
            & 4.8 [4.5, 5.1]
            & 16.7 [16.3, 17.1]
            & 11.8 [10.2, 13.6] \\
        &   & I-JEPA
            & 20.7 [20.2, 21.2]
            & 7.5 [7.1, 7.9]
            & 18.4 [17.9, 18.9]
            & 27.9 [26.0, 29.9] \\
        &   & DINO-v2
            & 19.8 [19.4, 20.2]
            & 6.2 [5.9, 6.5]
            & 17.1 [16.7, 17.5]
            & 13.9 [12.1, 15.8] \\
        &   & DINO-v3
            & 19.6 [19.2, 20.0]
            & 5.8 [5.5, 6.2]
            & 17.2 [16.8, 17.6]
            & 13.0 [11.3, 14.8] \\
        &   & RAD-DINO
            & 21.7 [21.3, 22.1]
            & 7.7 [7.3, 8.0]
            & 20.3 [19.9, 20.7]
            & \textbf{29.5 [28.0, 31.2]} \\
        &   & RadJEPA
            & \textbf{23.5 [23.2, 23.8]}
            & \textbf{8.6 [8.3, 8.9]}
            & \textbf{20.7 [20.4, 21.0]}
            & \underline{29.3 [27.9, 30.8]} \\

        \midrule

        \multirow{20}{*}{\textbf{IU-Xray}}

        & \multirow{5}{*}{MedLLaVA}
            & Base Encoder
            & 22.3 [21.9, 22.8]
            & 4.8 [4.5, 5.2]
            & 21.1 [20.6, 21.6]
            & 13.7 [11.8, 15.9] \\
        &   & I-JEPA
            & 26.2 [25.6, 26.8]
            & 8.4 [8.0, 8.8]
            & 25.2 [24.6, 25.8]
            & 22.5 [20.3, 24.8] \\
        &   & DINO-v2
            & 25.4 [24.9, 25.9]
            & 7.8 [7.4, 8.2]
            & 21.9 [21.3, 22.5]
            & 11.7 [9.8, 13.8] \\
        &   & DINO-v3
            & 24.8 [24.3, 25.3]
            & 7.5 [7.1, 7.9]
            & 22.2 [21.6, 22.9]
            & 11.3 [9.4, 13.4] \\
        &   & RAD-DINO
            & 26.7 [26.2, 27.1]
            & 8.6 [8.2, 9.0]
            & 25.8 [25.2, 26.3]
            & 23.9 [21.7, 26.2] \\
        &   & RadJEPA
            & \textbf{27.8 [27.4, 28.1]}
            & \textbf{9.1 [8.8, 9.4]}
            & \textbf{26.3 [25.9, 26.7]}
            & \textbf{25.8 [23.8, 27.8]} \\
        
        \cmidrule(lr){2-7}
        
        & \multirow{5}{*}{Qwen-2.5}
            & Base Encoder
            & 23.6 [23.1, 24.1]
            & 4.7 [4.3, 5.1]
            & 23.3 [22.8, 23.8]
            & 14.9 [12.8, 17.1] \\
        &   & I-JEPA
            & 29.0 [28.4, 29.7]
            & 9.3 [8.8, 9.8]
            & 25.9 [25.3, 26.6]
            & 22.5 [20.0, 25.1] \\
        &   & DINO-v2
            & 27.1 [26.6, 27.7]
            & 8.3 [7.9, 8.7]
            & 22.6 [22.0, 23.2]
            & 12.0 [10.1, 14.0] \\
        &   & DINO-v3
            & 26.7 [26.2, 27.2]
            & 7.9 [7.5, 8.4]
            & 22.3 [21.8, 22.9]
            & 11.7 [9.8, 13.8] \\
        &   & RAD-DINO
            & 28.4 [27.9, 28.9]
            & 9.6 [9.2, 10.0]
            & 26.1 [25.5, 26.7]
            & 25.8 [23.5, 28.2] \\
        &   & RadJEPA
            & \textbf{30.2 [29.8, 30.6]}
            & \textbf{10.1 [9.8, 10.4]}
            & \textbf{27.7 [27.2, 28.2]}
            & \textbf{26.9 [24.8, 29.0]} \\
        
        \cmidrule(lr){2-7}
        
        & \multirow{5}{*}{BLIP-2}
            & Base Encoder
            & 26.8 [26.3, 27.4]
            & 5.1 [4.7, 5.5]
            & 20.1 [19.5, 20.7]
            & 12.3 [10.4, 14.5] \\
        &   & I-JEPA
            & 27.5 [26.9, 28.1]
            & 9.1 [8.6, 9.6]
            & 24.9 [24.3, 25.5]
            & 23.6 [21.3, 26.0] \\
        &   & DINO-v2
            & 27.3 [26.8, 27.8]
            & 7.9 [7.5, 8.3]
            & 21.8 [21.2, 22.4]
            & 10.8 [8.9, 12.8] \\
        &   & DINO-v3
            & 27.0 [26.5, 27.5]
            & 7.8 [7.4, 8.3]
            & 22.0 [21.4, 22.6]
            & 10.2 [8.4, 12.2] \\
        &   & RAD-DINO
            & 28.2 [27.7, 28.7]
            & 9.0 [8.6, 9.4]
            & 26.6 [26.0, 27.2]
            & 25.3 [23.1, 27.6] \\
        &   & RadJEPA
            & \textbf{29.7 [29.3, 30.1]}
            & \textbf{9.4 [9.1, 9.7]}
            & \textbf{28.2 [27.7, 28.7]}
            & \textbf{25.9 [23.8, 28.0]} \\
        
        \cmidrule(lr){2-7}
        
        & \multirow{5}{*}{Phi-4}
            & Base Encoder
            & 20.2 [19.7, 20.7]
            & 4.2 [3.9, 4.6]
            & 18.4 [17.9, 18.9]
            & 9.6 [7.8, 11.5] \\
        &   & I-JEPA
            & 24.1 [23.5, 24.7]
            & 6.9 [6.5, 7.3]
            & 22.8 [22.2, 23.5]
            & 23.9 [21.6, 26.3] \\
        &   & DINO-v2
            & 22.9 [22.4, 23.4]
            & 6.1 [5.7, 6.5]
            & 20.7 [20.1, 21.2]
            & 10.0 [8.1, 12.0] \\
        &   & DINO-v3
            & 22.6 [22.1, 23.1]
            & 5.8 [5.4, 6.2]
            & 20.9 [20.3, 21.5]
            & 9.6 [7.8, 11.5] \\
        &   & RAD-DINO
            & 23.8 [23.3, 24.3]
            & 7.2 [6.8, 7.6]
            & 24.9 [24.3, 25.5]
            & \textbf{24.5 [22.2, 26.9]} \\
        &   & RadJEPA
            & \textbf{25.0 [24.6, 25.4]}
            & \textbf{8.3 [8.0, 8.6]}
            & \textbf{25.3 [24.8, 25.8]}
            & \underline{24.2 [22.0, 26.5]} \\

        \bottomrule
    \end{tabular}
    \end{adjustbox}

    \caption{Report-generation performance when the native vision encoder of each VLM is replaced by DINO-v2, I-JEPA, \raddino{}, or \radjepa{} (cf.\ Figure~\ref{fig:encoder_decoder}). \textit{Base Encoder} denotes the unmodified VLM. All experiments use the identical training recipe described in Section~\ref{sec:report_gen}. We report median and 95\% confidence intervals from $500$ bootstrap samples. For each VLM-metric cell, bold marks the best result among the encoder swaps; underline marks the second-best when the gap to the best is within the bootstrap interval (used only when \radjepa{} is the close runner-up to highlight that the gap is not statistically meaningful). The full statistical protocol is in Section~\ref{sec:significance}; complementary visual analysis is in Section~\ref{sec:appendix_vlm_portability}.}
    \label{tab:vlm_encoder_comparison}
\end{table*}

\begin{figure*}[!t]
    \centering
    \includegraphics[width=0.92\linewidth]{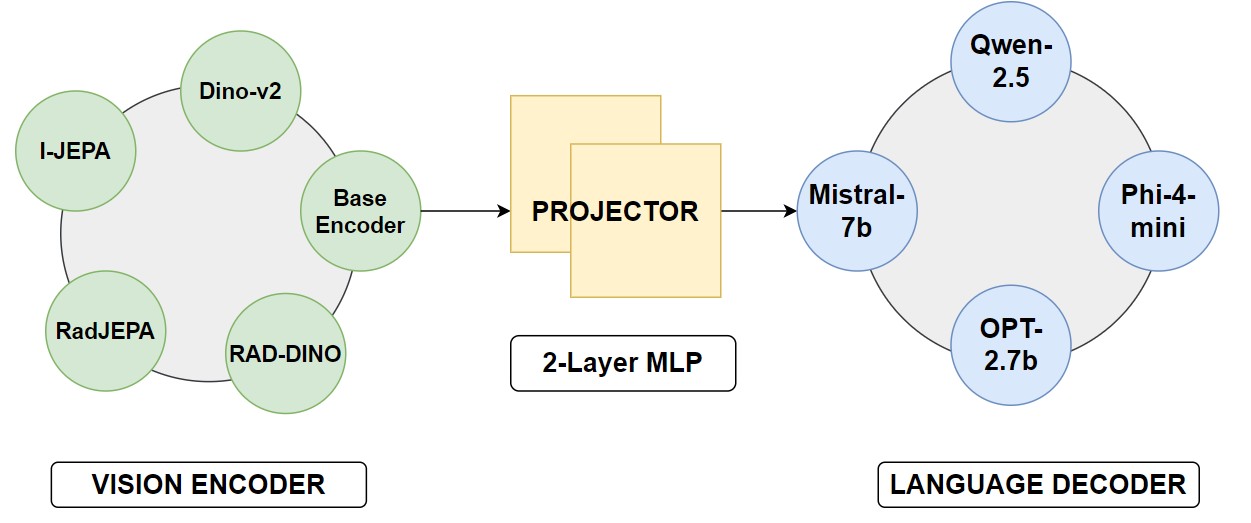}
    \caption{Encoder-swap protocol used in Table~\ref{tab:vlm_encoder_comparison}. The native vision encoder of each VLM is replaced by DINO-v2, I-JEPA, \raddino{}, or \radjepa{} while the projector and language decoder are kept unchanged. With four VLM backbones and four encoder swaps (plus the four base baselines), this gives $20$ configurations per dataset and $40$ runs in total across MIMIC-CXR and IU-Xray.}
    \label{fig:encoder_decoder}
\end{figure*}

\subsubsection*{Results analysis}

Table~\ref{tab:findings_generation} reports report generation results on MIMIC-CXR and IU-Xray using a fixed Vicuna-7B decoder with frozen image encoders. RadJEPA achieves the best performance across all metrics and datasets among image-only, predictive, and vision-language baselines. On MIMIC-CXR, it attains the highest ROUGE-L (\textbf{26.1}), BLEU-4 (\textbf{10.1}), RG\textsubscript{ER} (\textbf{23.8}), and Macro-F1-14 (\textbf{32.6}), outperforming \raddino{}, I-JEPA, and prior vision-language models. On IU-Xray, RadJEPA further improves ROUGE-L by \textbf{+1.7}, RG\textsubscript{ER} by \textbf{+0.9}, and Macro-F1-14 by \textbf{+1.3} over the strongest non-JEPA baseline, indicating robust out-of-distribution generalization. These gains are achieved despite operating at lower input resolution than contrastive baselines such as \raddino{}. Under a controlled MIMIC-only pretraining setting, \radjepacontrol{} matches or exceeds architecture-matched baselines, while full RadJEPA yields additional improvements. The same ranking holds under the LLM-based GREEN clinical-correctness metric~\citep{ostmeier2024green} (Appendix~\ref{sec:appendix_green}).

\subsection{Extended Comparison with RAD-DINO}
\label{sec:encoder_decoder}
The objective of our work is to compare the performance of image-text paired vision encoders and DINO-style self-distillation methods against Joint Embedding Predictive Architectures (JEPA). In Table \ref{tab:findings_generation}, we compare both image-text paired vision encoders and DINO-style self-distillation encoders (DINO-v2 and RAD-DINO). In this subsection, we further extend the comparison of RadJEPA with a particular focus on RAD-DINO and its foundational model DINO-v2 to support our claim that, under similar settings, Joint Embedding Predictive Architectures demonstrate superior performance over DINO-style view-augmented self-supervised models for radiology chest X-ray images. We do not include comparisons with other SSL methods such as SimCLR, as no large-scale chest X-ray pretrained SimCLR \cite{chen2020simple} models exist, falling outside the scope of this study.

\paragraph{Baseline Models: }
We evaluate four widely used Vision-Language Models (VLMs): MedLLaVA~\cite{li2023llava}, Phi-4~\cite{abdin2024phi}, Qwen-2.5~\cite{yang2025qwen3}, and BLIP-2~\cite{li2023blip}. We choose MedLLaVA instead of the original LLaVA, as LLaVA employs the Vicuna-7B (v1.5) decoder, extensively evaluated in Table~\ref{tab:findings_generation}. Public checkpoint URLs for every backbone, decoder, and image encoder used in this paper are listed in Appendix~\ref{sec:appendix_checkpoints}.

For all experiments, each VLM is trained on the MIMIC-CXR dataset under the same experimental setting described in Section~\ref{sec:report_gen}, including identical dataset splits and prompting strategies. We first evaluate each VLM without modifying its original architecture, and report these results in Table~\ref{tab:vlm_encoder_comparison} under the \textit{Base Encoder} setting. The corresponding language decoders are Mistral-7B for MedLLaVA, Qwen2.5-7B for Qwen-2.5, OPT-2.7B~\cite{zhang2022opt} for BLIP-2, and Phi-4-mini for Phi-4. We then replace the original vision encoder of each VLM with DINO-v2, I-JEPA, \raddino{}, and \radjepa{} (cf.\ Figure~\ref{fig:encoder_decoder}), while keeping the remaining architecture unchanged.

\paragraph{Experimental Setup: }We follow the standard LLaVA-style architecture, where embeddings from the frozen vision encoder are projected through a two-layer MLP to match the embedding dimension required by the language decoder. Since both RadJEPA and RAD-DINO are based on the ViT-B architecture, we restrict our experiments to language models with up to 7B parameters. Except for the vision encoder, all remaining components, including the projection layer and the LLM decoder, are fully trainable during fine-tuning. The complete results are reported in Table \ref{tab:vlm_encoder_comparison}.

\begin{figure*}[t]
    \centering
    \includegraphics[width=\linewidth]{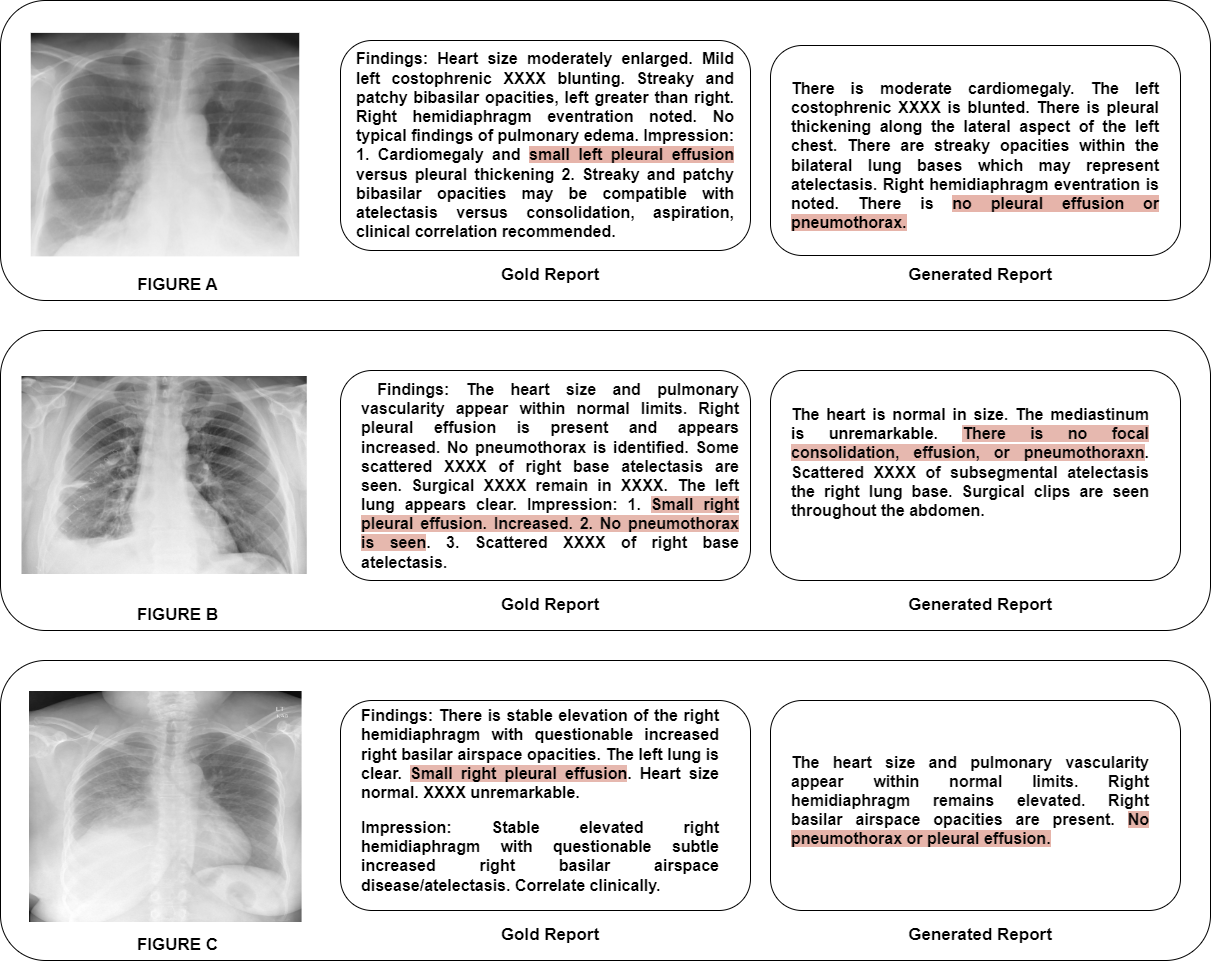}
\caption{Examples of error in RadJEPA-generated reports, where the model predicts the absence of pleural effusion and pneumothorax despite positive findings in the ground-truth report.
}
    \label{fig:error_analysis}
\end{figure*}

\paragraph{Result Analysis: }\radjepa{} achieves the best overall performance in almost every cell of Table~\ref{tab:vlm_encoder_comparison}, with \raddino{} as the typical runner-up. On MIMIC-CXR, the relative gains over \raddino{} are approximately $\textbf{2.8\%}$, $\textbf{8.1\%}$, $\textbf{2.7\%}$, and $\textbf{4.6\%}$ on the four metrics, averaged across the four VLM backbones; on IU-Xray the same gains are $\textbf{4.0\%}$, $\textbf{5.7\%}$, $\textbf{4.8\%}$, and $\textbf{3.2\%}$. The only exception is Macro-F1-14 with the Phi-4 backbone, where \raddino{} is ahead by $\textbf{0.7}$--$\textbf{1.2\%}$ across the two datasets, likely reflecting the smaller (${\approx}3.5$B) Phi-4-mini decoder versus the ${\approx}7$B Mistral/Qwen/Vicuna decoders used by the other VLMs. DINO-v3 underperforms both RadJEPA across all downstream tasks and DINO-v2 across many downstream tasks. This is likely influenced by the publicly available DINO-v3 model using a ViT-B/16 backbone\footnote{\href{https://huggingface.co/facebook/dinov3-vitb16-pretrain-lvd1689m}{\texttt{facebook/dinov3-vitb16-pretrain-lvd1689m}}} with 86M parameters, comparable in scale to RAD-DINO and RadJEPA, whereas DINO-v2 uses a substantially larger ViT-G/14 backbone with 1.1B parameters.


\section{Error Analysis}

\begin{table}[t]
    \centering
    \small
    \begin{tabular}{c|cc}
        \toprule
         & $\mathcal{N}^{+}$ & $\mathcal{N}^{-}$ \\
         & 7,667 & 195,425 \\
        \midrule
        $\mathcal{B}^{+}$ & 3,552 (0.463) & 30,988 (0.159) \\
        $\mathcal{B}^{-}$ & 1,345 (0.175) & 153,822 (0.787) \\
        \bottomrule
    \end{tabular}
    \caption{
    Data distribution bias of Pleural Effusion ($\mathcal{B}$) conditioned on Pneumothorax ($\mathcal{N}$), i.e., $P(\mathcal{B}^{*}|\mathcal{N}^{*})$, in the MIMIC-CXR dataset. `+' and `-' denote abnormal and normal conditions, respectively. Integer values represent the number of samples, while the decimal values in parentheses denote conditional probabilities.
    }
    \label{tab:bias_example}
\end{table}

The largest dataset used during RadJEPA pretraining is MIMIC-CXR-JPG, contributing approximately \textbf{36\%} of the total pretraining data. Chest X-ray datasets often suffer from dataset-specific biases, as they are collected under constrained clinical settings \cite{jones2024causal}. For example, MIMIC-CXR was collected from Beth Israel Deaconess Medical Center, primarily from ICU patients, making the dataset vulnerable to strong correlations between findings that may not necessarily be causally related. One such example is the correlation between \textit{Pleural Effusion} and \textit{Pneumothorax} \cite{song2024rethinking}; cf.\ Table \ref{tab:bias_example}.

During qualitative analysis, we observed multiple bias behaviours in \radjepa{}-generated reports where the model simultaneously predicted the absence of both pleural effusion and pneumothorax, even when one of these findings was clearly present in the ground-truth report. To quantify this, we manually re-examined the $200$ cases in our human-annotated sample (Section~\ref{sec:appendix_iaa}) whose ground-truth reports mention pleural effusion or pneumothorax. Within this subset, \radjepa{} produced $33$ such joint-absence errors against $37$ for \raddino{} under the same decoder. The reduction is modest and we surface it here so that future work is aware of the failure mode rather than treat it as solved. All examples shown in Figure \ref{fig:error_analysis} belong to the IU-Xray dataset and use the Vicuna-7B decoder. In the first example (Figure A), the original report mentions a small left pleural effusion, whereas the RadJEPA-generated report incorrectly states that neither pleural effusion nor pneumothorax is present. Similarly, in the second example (Figure B), a right pleural effusion is present in the ground-truth report, but the generated report again predicts the absence of both findings. In the third example (Figure C), a similar scenario was observed. Beyond the representative failures in Figure~\ref{fig:error_analysis}, Appendix~\ref{app:qualitative_errors} provides a broader qualitative analysis of the human-reviewed reports, including false-positive findings, hallucinated clinical content, missed subtle abnormalities, and examples from low-frequency disease categories.

\paragraph{Inter-Annotator Agreement:}
Two clinical annotators and three LLM judges (aggregated by per-metric majority vote) independently graded a stratified $500$-report sample drawn equally from IU-Xray and MIMIC-CXR. Inter-human quadratic-weighted Cohen's $\kappa$ is $0.91$ on the 0--5 score and $0.83$ on the binary acceptability flag~\citep{landis1977measurement}; the LLM vote agrees with the humans at $\kappa{=}0.84$ and $0.81$, and the five-rater Krippendorff's $\alpha$ is $0.77$. All values fall in the substantial-to-almost-perfect range on the Landis-Koch scale. Full statistics are in Appendix~\ref{sec:appendix_iaa}.

\section{Conclusion \& Future Works}

We introduced \radjepa{}, a language-supervision-free Joint Embedding Predictive encoder for chest X-rays, pretrained on ${\approx}840$K unlabeled images. On report generation, \radjepa{} matches or exceeds the strongest image-only and vision-language baselines across four VLM backbones, two datasets, and four metrics, and shows consistent improvements on classification and segmentation. The radiologist study demonstrates high agreement in judgments of RadJEPA generated reports, but it does not constitute a paired clinical comparison with baselines. Future work should evaluate broader institutions and views, perform paired radiologist comparisons, and study bias and collapse mitigation in JEPA-style radiology encoders.


\section{Limitation}
\label{sec:limitation}
In this section, we briefly discuss the limitations of our work arising from both the characteristics of the datasets and the JEPA-style architecture used in this study.

\subsection{Dataset Limitations}

Although \radjepa{} is pretrained on approximately $840$K chest X-ray images collected from five large open-source datasets, these datasets still contain several inherent limitations. Most datasets, such as MIMIC-CXR and CheXpert, are collected from specific hospitals and clinical environments, introducing demographic, acquisition, and disease-distribution biases that may not generalize across institutions. Additionally, several instances contain strong label co-occurrence correlations (e.g., pleural effusion and pneumothorax) \cite{song2024rethinking}, which may encourage shortcut learning rather than true causal understanding \cite{jones2024causal}. The datasets also vary considerably in image quality, acquisition protocols, and reporting styles, introducing domain shifts during pretraining. Similarly, the downstream evaluation datasets also contain limitations. Chest X-ray benchmarking datasets are relatively small and may not fully reflect the diversity and complexity of real-world clinical settings, a limitation also commonly observed in multimodal evaluation benchmarks outside the medical domain~\cite{khan2025early}. Furthermore, evaluations are restricted to frontal chest X-rays and may not fully capture the diversity of real-world radiology settings, thereby limiting broader clinical generalization.

Similarly, the downstream evaluation datasets also contain limitations. Chest X-ray evaluation benchmarks are relatively small and may not fully reflect the diversity and complexity of real-world clinical settings, a limitation also observed in other multimodal evaluation benchmarks~\cite{khan2025early}. This make performance estimates noisy and restrict conclusions about broader generalization.

\subsection{Architectural Limitations}

One limitation of JEPA-style architectures is that collapse prevention is not explicitly enforced at the objective level, but instead relies on architectural design choices such as teacher-student asymmetry and EMA-based updates. As discussed in prior theoretical analyses comparing PMAX (similar to JEPA-style architectures) with PMIN, PMAX type of architecture (JEPA) does not directly constrain internal redundancy or enforce entropy balancing within representations, allowing correlated or redundant latent features to persist. In practice, this may lead to occasional representation collapse. Moreover, JEPA-style models remain sensitive to factors such as masking strategy, teacher momentum, and dataset bias, all of which can affect training stability and representation quality.

\section{Ethical Considerations}

All datasets used in this work are publicly available open-source datasets obtained from their respective official sources and appropriately cited throughout the paper. The datasets are de-identified and do not contain personally identifiable patient information. Furthermore, all pretrained models, VLM backbones, and baseline methods used in this study are open-source and used in accordance with their respective licenses and research guidelines. As discussed in the Limitation section (cf.\ Section \ref{sec:limitation}), both dataset bias and architectural bias may affect the learned representations of RadJEPA. Despite efforts to mitigate such biases, it remains challenging to eliminate biased or discriminatory patterns from the learned representations and generated outputs. AI assistants were used for writing support and
sentence polishing.

\section*{Acknowledgement}

We gratefully acknowledge the Computation for Indian Language Technology (CFILT) - IIT Bombay, Koita Centre for Digital Health - IIT Bombay, IIT Bombay Trust Lab, and the Department of Computer Science and Engineering - IIT Bombay, for their support, resources, and funding that enabled this work. We are sincerely grateful for their generous support. We also remember with deep respect and gratitude Prof. Pushpak Bhattacharyya, whose mentorship, encouragement, and invaluable contributions continue to inspire this work.


\section{Evaluation on Downstream Tasks}
\label{sec:class_seg}

\begin{table*}[!t]
    \centering
    \footnotesize
    \begin{adjustbox}{width=\textwidth}
    \begin{tabular}{
        l
        *{7}{>{\centering\arraybackslash}p{1.35cm}}
        *{2}{>{\centering\arraybackslash}p{1.35cm}}
    }
        \toprule
        & \multicolumn{7}{c}{VinDr-CXR (AUPRC)} & \multicolumn{2}{c}{RSNA} \\
        \cmidrule(lr){2-8} \cmidrule(lr){9-10}
        Model
        & LO & CM & PL-T & AE & PF & PE & Agg.
        & AUPRC & AUC \\
        \midrule
        CLIP@224
            & 9.7 $\pm$ 0.4
            & 42.6 $\pm$ 0.2
            & 18.8 $\pm$ 0.4
            & 30.0 $\pm$ 0.5
            & 24.1 $\pm$ 0.4
            & 21.8 $\pm$ 0.4
            & 23.8
            & 60.1 $\pm$ 2.0
            & 83.7 $\pm$ 0.7 \\
        
        CLIP@336
            & 9.1 $\pm$ 0.1
            & 46.1 $\pm$ 0.2
            & 18.5 $\pm$ 0.2
            & 29.0 $\pm$ 0.3
            & 22.8 $\pm$ 0.3
            & 18.6 $\pm$ 0.3
            & 23.4
            & 60.0 $\pm$ 1.7
            & 84.2 $\pm$ 0.4\\
        
        \midrule
        
        BioViL-T
            & 12.7 $\pm$ 0.1
            & 51.4 $\pm$ 0.5
            & 24.6 $\pm$ 0.2
            & 22.3 $\pm$ 0.1
            & 30.5 $\pm$ 0.1
            & 52.2 $\pm$ 0.4
            & 32.4
            & 66.8 $\pm$ 1.5
            & 86.9 $\pm$ 0.5\\
        
        BiomedCLIP
            & 10.0 $\pm$ 0.3
            & 58.5 $\pm$ 0.8
            & 24.4 $\pm$ 0.5
            & 36.2 $\pm$ 0.2
            & 32.0 $\pm$ 0.6
            & 54.1 $\pm$ 0.6
            & 35.9
            & 68.4 $\pm$ 1.7
            & 87.5 $\pm$ 0.4\\
        
        \midrule
        
        CheXzero
            & 11.1 $\pm$ 0.6
            & 74.4 $\pm$ 0.2
            & 25.1 $\pm$ 0.3
            & 42.9 $\pm$ 0.2
            & 33.1 $\pm$ 0.4
            & 60.2 $\pm$ 0.5
            & 40.0
            & 68.9 $\pm$ 1.9
            & 87.9 $\pm$ 0.4\\
        
        MRM
            & 12.2 $\pm$ 0.3
            & \textbf{79.7 $\pm$ 0.4}
            & 35.8 $\pm$ 0.3
            & 47.7 $\pm$ 0.2
            & 47.1 $\pm$ 0.4
            & 77.2 $\pm$ 0.3
            & 51.3
            & 71.4 $\pm$ 1.5
            & 89.0 $\pm$ 0.5\\
        \midrule
        DINO-v2
            & 12.6 $\pm$ 0.4
            & 67.3 $\pm$ 0.5
            & 31.7 $\pm$ 0.4
            & 38.9 $\pm$ 0.3
            & 53.5 $\pm$ 0.5
            & 69.8 $\pm$ 0.4
            & 45.6
            & 67.6 $\pm$ 1.8
            & 85.9 $\pm$ 0.6 \\
        DINO-v3
            & 11.8 $\pm$ 0.5
            & 66.9 $\pm$ 0.4
            & 29.7 $\pm$ 0.3
            & 35.4 $\pm$ 0.4
            & 52.0 $\pm$ 0.5
            & 66.5 $\pm$ 0.5
            & 43.7
            & 65.2 $\pm$ 1.6
            & 82.8 $\pm$ 0.4 \\

        \raddino~\cite{perez2025exploring}
            & 14.2 $\pm$ 0.3
            & 68.4 $\pm$ 0.4
            & 35.8 $\pm$ 0.5
            & 48.5 $\pm$ 0.3
            & 57.2 $\pm$ 0.4
            & 74.9 $\pm$ 0.4
            & 49.8
            & 69.1 $\pm$ 1.7
            & 86.8 $\pm$ 0.5 \\
        CheXWorld~\cite{yue2025chexworld}
            & 14.5 $\pm$ 0.3
            & 69.1 $\pm$ 0.4
            & 36.1 $\pm$ 0.5
            & 44.1 $\pm$ 0.3
            & 58.7 $\pm$ 0.4
            & \textbf{76.9 $\pm$ 0.4}
            & 49.9
            & 70.4 $\pm$ 1.7
            & 88.0 $\pm$ 0.5 \\
        \midrule
        I-JEPA
            & 13.6 $\pm$ 0.3
        & 68.3 $\pm$ 0.6
        & 32.8 $\pm$ 0.7
        & 41.9 $\pm$ 0.5
        & 55.0 $\pm$ 0.5
        & 70.5 $\pm$ 0.6
        & 47.0
        & 68.2 $\pm$ 2.0
        & 86.4 $\pm$ 0.8 \\

        \rowcolor{gray!15}
        RadJEPA 
            & \textbf{19.2 $\pm$ 0.2}
            & {72.6 $\pm$ 0.3}
            & \textbf{42.2 $\pm$ 0.4}
            & \textbf{58.6 $\pm$ 0.2}
            & \textbf{63.9 $\pm$ 0.3}
            & 74.4 $\pm$ 0.4
            & \textbf{55.2}
            & \textbf{72.7 $\pm$ 1.5}
            & \textbf{89.2 $\pm$ 0.4} \\
        \bottomrule
    \end{tabular}
    \end{adjustbox}
    \caption{
    Image classification performance with linear probing on VinDr-CXR and RSNA-Pneumonia. AUPRC denotes area under the precision-recall curve and AUC denotes area under the ROC curve. Agg.\ denotes mean AUPRC over the six VinDr-CXR classes: LO (Lung Opacity), CM (Cardiomegaly), PL-T (Pleural Thickening), AE (Aortic Enlargement), PF (Pulmonary Fibrosis), and PE (Pleural Effusion). The \raddino{} and CheXWorld rows are reproduced from publicly released checkpoints under our evaluation pipeline. VinDr-CXR uses $15{,}000$/$3{,}000$ train/test splits; RSNA results are reported on the test set ($5{,}337$ images). Numbers in bold are best per column. Results are averaged over five random seeds; the full statistical protocol is in Section~\ref{sec:significance}.
    }
    \label{tab:classification_vindr_rsna}
\end{table*}

\subsection{Image Classification}

\subsubsection{Datasets and Implementation Details}
We evaluate on VinDr-CXR~\cite{nguyen2022vindr} and RSNA-Pneumonia \footnote{\url{https://www.kaggle.com/c/rsna-pneumonia-detection-challenge}}. For VinDr-CXR (diverse thoracic findings), we use six findings with a 15{,}000/3{,}000 train/test split by subject. For RSNA-Pneumonia (26{,}684 images; acute lung opacities), we adopt a 60/20/20 train/validation/test split by subject. All splits are performed by subject identifier. Images are converted to single-channel grayscale, center-cropped, and resized. Training augmentations include random horizontal flipping, affine transformations, random cropping, color jittering, and additive Gaussian noise. Classification is performed via linear probing, where a sigmoid-activated linear classifier is trained on frozen 768-dimensional global embeddings using binary cross-entropy loss. Optimization uses AdamW with a learning rate of $5\times10^{-5}$ and cosine scheduling. Experiments are conducted on {3 NVIDIA A6000 GPUs} with batch size 32 per GPU (total 96) for 100 epochs. Results are reported as mean of AUPRC over 5-fold cross-validation.

\subsubsection{Results analysis}

Table~\ref{tab:classification_vindr_rsna} reports results on VinDr-CXR and RSNA-Pneumonia. RadJEPA outperforms all image-only and vision-language baselines across both datasets. On VinDr-CXR, it achieves the highest mean AUPRC (Agg.) of \textbf{55.2}, surpassing \raddino{} (52.8), I-JEPA (47.0), and MRM (51.3). On RSNA-Pneumonia, it attains the best AUPRC (\textbf{72.7}) and AUROC (\textbf{89.2}). Notably, RadJEPA yields strong gains on subtle findings (PL-T, AE, PF), improving PF by \textbf{+4.5} over \raddino{} and \textbf{+8.9} over I-JEPA. Compared to I-JEPA, consistent improvements are observed across all VinDr-CXR classes. Despite using a ViT-B/14 backbone (as shown in Table~\ref{tab:all_model_details}) with only \textbf{86M} parameters (vs. 1.1B for DINO-v2 and 0.6B for I-JEPA), RadJEPA achieves superior downstream performance.


\begin{table*}[!ht]
    \centering
    \footnotesize
    \begin{adjustbox}{width=\textwidth}
    \begin{tabular}{
        llrr
        *{3}{>{\centering\arraybackslash}p{1.6cm}}
    }
        \toprule
        \textbf{Encoder} & \textbf{Decoder} & \textbf{\# Features} & \textbf{\# Params}
        & \textbf{Lungs} & \textbf{Lung zones} & \textbf{Ribs} \\
        \midrule
        NN-UNet~\cite{isensee2018nnu}
            & Unet & -- & 17.9 M
            & 98.0 (1.1)
            & 92.6 (10.2)
            & 86.2 (2.8) \\
        
        EfficientNet-B6~\cite{tan2019efficientnet}
            & Unet & -- & 45.9 M
            & 98.3 (1.1)
            & 92.7 (10.1)
            & 88.9 (2.6) \\
        
        \midrule
        
        BioViL-T~\cite{bannur2023learning}
            & Linear & 2048 & 2049
            & 83.2 (3.2)
            & 69.4 (9.0)
            & 59.1 (4.7) \\
        
        BiomedCLIP~\cite{zhang2023biomedclip}
            & Linear & 768 & 769
            & 90.4 (2.6)
            & 76.2 (10.2)
            & 67.4 (4.5) \\
        
        CheXzero~\cite{tiu2022expert}
            & Linear & 768 & 769
            & 84.0 (3.4)
            & 68.3 (9.1)
            & 62.0 (3.3) \\
        
        \raddino~\cite{perez2025exploring}
            & Linear & 768 & 769
            & 95.9 (1.5)
            & 85.7 (9.8)
            & 73.4 (3.6) \\
        DINO-v2
            & Linear & 768 & 769
            & 90.5 (2.7)
            & 77.9 (9.8)
            & 70.2 (4.2) \\
        DINO-v3
            & Linear & 768 & 769
            & 89.1 (1.4)
            & 77.4 (9.3)
            & 69.7 (3.1) \\
        I-JEPA
            & Linear & 768 & 769
            & 91.1 (2.4)
            & 81.6 (9.4)
            & 71.8 (4.0) \\
        RadJEPA
            & Linear & 768 & 769
            & 96.2 (1.4)
            & 86.5 (9.2)
            & 73.9 (3.4) \\
        \midrule
        DINO-v2
            & ViTDet & $4 \times 768$ & 24.8 M
            & 96.1 (1.4)
            & 88.3 (9.3)
            & 79.4 (3.1) \\
        DINO-v2
            & UPerNet & $4 \times 768$ & 39.3 M
            & 96.4 (1.3)
            & 88.7 (9.1)
            & 80.1 (2.9) \\
        DINO-v3
            & ViTDet & $4 \times 768$ & 24.8 M
            & 96.0 (1.2)
            & 88.4 (9.0)
            & 79.6 (2.6) \\

        DINO-v3
            & UPerNet & $4 \times 768$ & 39.3 M
            & 95.9 (0.9)
            & 89.2 (8.8)
            & 80.3 (2.2) \\
        \raddino~\cite{perez2025exploring}
            & ViTDet & $4 \times 768$ & 24.8 M
            & 97.8 (1.2)
            & 90.8 (9.9)
            & 83.7 (2.8) \\
        \raddino~\cite{perez2025exploring}
            & UPerNet & $4 \times 768$ & 39.3 M
            & 98.1 (1.1)
            & 91.5 (9.8)
            & 84.8 (2.7) \\
        CheXWorld~\cite{yue2025chexworld}
            & ViTDet & $4 \times 768$ & 24.8 M
            & 97.6 (1.3)
            & 90.3 (10.0)
            & 83.1 (2.9) \\
        CheXWorld~\cite{yue2025chexworld}
            & UPerNet & $4 \times 768$ & 39.3 M
            & 97.9 (1.2)
            & 91.0 (9.9)
            & 84.2 (2.8) \\
        \midrule
        I-JEPA
            & ViTDet & $4 \times 768$ & 24.8 M
            & 96.6 (1.3)
            & 90.2 (9.7)
            & 82.4 (2.9) \\
        I-JEPA
            & UPerNet & $4 \times 768$ & 39.3 M
            & 96.9 (1.2)
            & 91.0 (9.5)
            & 83.2 (2.8) \\
        RadJEPA
            & ViTDet & $4 \times 768$ & 24.8 M
            & 98.0 (1.1)
            & 92.9 (9.0)
            & 88.0 (2.3) \\
        \rowcolor{gray!15}
        RadJEPA
            & UPerNet & $4 \times 768$ & 39.3 M
            & \textbf{98.3 (1.0)}
            & \textbf{93.7 (8.8)}
            & \textbf{89.6 (2.1)} \\
        \bottomrule
    \end{tabular}
    \end{adjustbox}
    \caption{
    Semantic segmentation results obtained with linear, ViTDet~\cite{li2022exploring}, and UPerNet~\cite{xiao2018unified} decoders on frozen backbone encoders. Dice scores are reported as mean (standard deviation in parentheses). Lungs denotes left/right lung segmentation, Lung zones the six anatomical lung regions, and Ribs the 20 individual ribs. End-to-end U-Net models indicate an upper-bound reference. \raddino{} and CheXWorld rows are reproduced from publicly released checkpoints under our evaluation pipeline.
    }
    \label{tab:segmentation_benchmarks}
\end{table*}

\subsection{Semantic Segmentation}

\subsubsection{Datasets and Implementation Details}
{Lung and Lung Zone Segmentation} masks are obtained using Chest ImaGenome bounding boxes for six regions (left/right upper, middle, lower)~\cite{wu2021chest}. This yields 1{,}138 images from 1{,}138 subjects. For \textbf{Rib Segmentation}, we use the VinDR-RibCXR dataset~\cite{nguyen2021vindr}, which provides expert annotations for 20 ribs (L1--L10, R1--R10) across 245 subjects, following the official train/test splits. We evaluate RadJEPA for semantic segmentation using frozen backbone encoders with task-specific decoder heads. Experiments are conducted on a single node with {4 NVIDIA A6000 GPUs} (batch size 20 per GPU; total 80). Models are optimized with Adam ($5\times10^{-4}$) and cosine scheduling for 100 epochs. Images are center-cropped and resized; training augmentations include random horizontal flipping (except for left-right lung and lung zone tasks), random affine and elastic transformations, brightness/contrast jittering, and random gamma adjustments. Image intensities are normalized using statistics computed from all MIMIC-CXR images~\cite{johnson2019mimic}. Data are split 70/15/15 by subject, and evaluation is performed on the test set using the checkpoint with lowest validation loss (rib segmentation follows the official 196/49 train/test split). Results are reported as mean of Dice score over 5-fold cross-validation.

\subsubsection{Results analysis}
Table~\ref{tab:segmentation_benchmarks} reports semantic segmentation performance. Across all segmentation tasks, \textit{RadJEPA consistently outperforms competing image-only and vision-language pretrained backbones} under identical decoder configurations. With a UPerNet decoder, RadJEPA achieves the highest Dice scores on lung (\textbf{98.3}), lung zone (\textbf{93.7}), and rib (\textbf{89.6}) segmentation, surpassing \raddino{} and I-JEPA across all three tasks. Notably, RadJEPA shows the largest gains on structurally complex targets such as lung zones and ribs, improving rib segmentation by \textbf{+4.8 Dice} over \raddino{} and \textbf{+6.4 Dice} over I-JEPA. Importantly, all predictive and contrastive baselines share the same decoder.

\section{Model Checkpoints}
\label{sec:appendix_checkpoints}

Table~\ref{tab:model_checkpoints} lists the public Hugging Face checkpoint used for every backbone, decoder, and image encoder evaluated in this paper. 
\begin{table}[!ht]
\centering
\small
\renewcommand{\arraystretch}{1.15}
\begin{tabular}{ll}
\toprule
\textbf{Model} & \textbf{Source} \\
\midrule
\multicolumn{2}{l}{Vision-Language backbones (Section~\ref{sec:encoder_decoder})} \\
MedLLaVA       & \href{https://huggingface.co/microsoft/llava-med-v1.5-mistral-7b}{HF} \\
Phi-4          & \href{https://huggingface.co/microsoft/Phi-4-multimodal-instruct}{HF} \\
Qwen-2.5       & \href{https://huggingface.co/Qwen/Qwen2.5-7B-Instruct}{HF} \\
BLIP-2         & \href{https://huggingface.co/Salesforce/blip2-opt-2.7b}{HF} \\
\midrule
\multicolumn{2}{l}{Language decoders} \\
Vicuna-7B (v1.5) & \href{https://huggingface.co/lmsys/vicuna-7b-v1.5}{HF} \\
Mistral-7B       & \href{https://huggingface.co/mistralai/Mistral-7B-Instruct-v0.3}{HF} \\
Qwen2.5-7B       & \href{https://huggingface.co/Qwen/Qwen2.5-7B-Instruct}{HF} \\
OPT-2.7B         & \href{https://huggingface.co/facebook/opt-2.7b}{HF} \\
Phi-4-mini       & \href{https://huggingface.co/microsoft/Phi-4-mini-instruct}{HF} \\
\midrule
\multicolumn{2}{l}{Image encoders evaluated in Tables~\ref{tab:findings_generation} and \ref{tab:vlm_encoder_comparison}} \\
DINO-v2          & \href{https://huggingface.co/facebook/dinov2-giant}{HF} \\
I-JEPA           & \href{https://huggingface.co/facebook/ijepa_vith14_1k}{HF} \\
\raddino{}       & \href{https://huggingface.co/microsoft/rad-dino}{HF} \\
CheXWorld        & \cite{yue2025chexworld} \\
\midrule
\multicolumn{2}{l}{Evaluation models} \\
CheXbert         & \cite{smit2020chexbert} \\
RadGraph         & \cite{jain2021radgraphextractingclinicalentities} \\
GREEN            & \href{https://huggingface.co/StanfordAIMI/GREEN-radllama2-7b}{HF} \\
\bottomrule
\end{tabular}
\caption{Hugging Face checkpoints for every model referenced in the main paper. ``HF'' is a hyperlink to the repository.}
\label{tab:model_checkpoints}
\end{table}

\section{Statistical Evaluation Protocol}
\label{sec:significance}

\paragraph{Motivation:}
A common concern in deep learning evaluation is the choice of statistical significance analysis under limited experimental runs. Classical paired significance tests such as paired t-tests assume approximate Gaussianity of the sampled differences, while non-parametric alternatives such as the Wilcoxon signed-rank test require a sufficient number of samples to reliably establish significance. In our setting, each experiment requires large-scale training on chest X-ray images and multiple VLM backbones, making a substantially larger number of runs computationally expensive and practically infeasible. Therefore, instead of relying on potentially unstable significance tests with only five runs, we adopt the same statistical evaluation strategy used in RAD-DINO for fair and consistent comparison.

\paragraph{Report Generation.}
For radiology report generation (cf.\ Tables \ref{tab:findings_generation} and \ref{tab:vlm_encoder_comparison}), following RAD-DINO, we report the median and 95\% confidence intervals obtained from 500 bootstrap samples. Across most evaluation settings, RadJEPA achieves higher median performance along with narrower confidence intervals compared to RAD-DINO, suggesting improved robustness and reduced evaluation uncertainty.

\paragraph{Classification and Segmentation.}
For disease classification and semantic segmentation tasks (cf.\ Tables \ref{tab:classification_vindr_rsna} and \ref{tab:segmentation_benchmarks}), we report the mean and standard deviation across five runs with different random seeds. Across most settings, RadJEPA achieves higher mean performance while also exhibiting lower standard deviation compared to RAD-DINO, indicating more stable and consistent representations.

\section{CheXWorld Overview}
\label{sec:CheXWorld}

CheXWorld \cite{yue2025chexworld} is a recently proposed image-only chest X-ray representation learning framework based on an architecture similar to I-JEPA. The model uses a ViT-B/14 backbone with approximately 86M parameters and is pretrained on around 448K frontal chest X-ray images collected from multiple datasets at a resolution of $224\times224$. Unlike RadJEPA, CheXWorld does not utilize lateral images during pretraining and is trained on approximately 31\% fewer images. The pretrained weights are publicly available, which enabled us to reproduce and compare the model across all downstream tasks considered in our work.

We observed trends similar to those reported in the original CVPR 2025 paper: CheXWorld generally performs competitively with RAD-DINO, but in most settings slightly underperforms RAD-DINO and consistently remains below RadJEPA. One possible reason is the comparatively smaller pretraining dataset and lower training compute budget, which may impose limitations on optimization choices such as batch size and large-scale training stability. Furthermore, the original CheXWorld work did not evaluate report generation performance. In fact, several recent chest X-ray representation learning works, including X-Win \cite{yang2025x, jalan-etal-2026-medbench, Banerjee2026.07.23.740366}, primarily focus on classification and segmentation benchmarks without extensive report generation analysis.

Although the primary objective of our work is radiology report generation, we additionally perform extensive classification and segmentation experiments to provide a more robust and comprehensive comparison across existing representation learning frameworks. Following prior radiology benchmarking practices (including RAD-DINO from Microsoft), we primarily report AUPRC instead of AUROC due to the severe class imbalance and non-mutually exclusive nature of chest X-ray disease labels. Nevertheless, for completeness, we also report reproduced AUROC comparisons for CheXWorld, RAD-DINO, and RadJEPA in Table \ref{tab:chexworld_auroc}.

\begin{figure*}[!t]
    \centering
    \includegraphics[width=\textwidth]{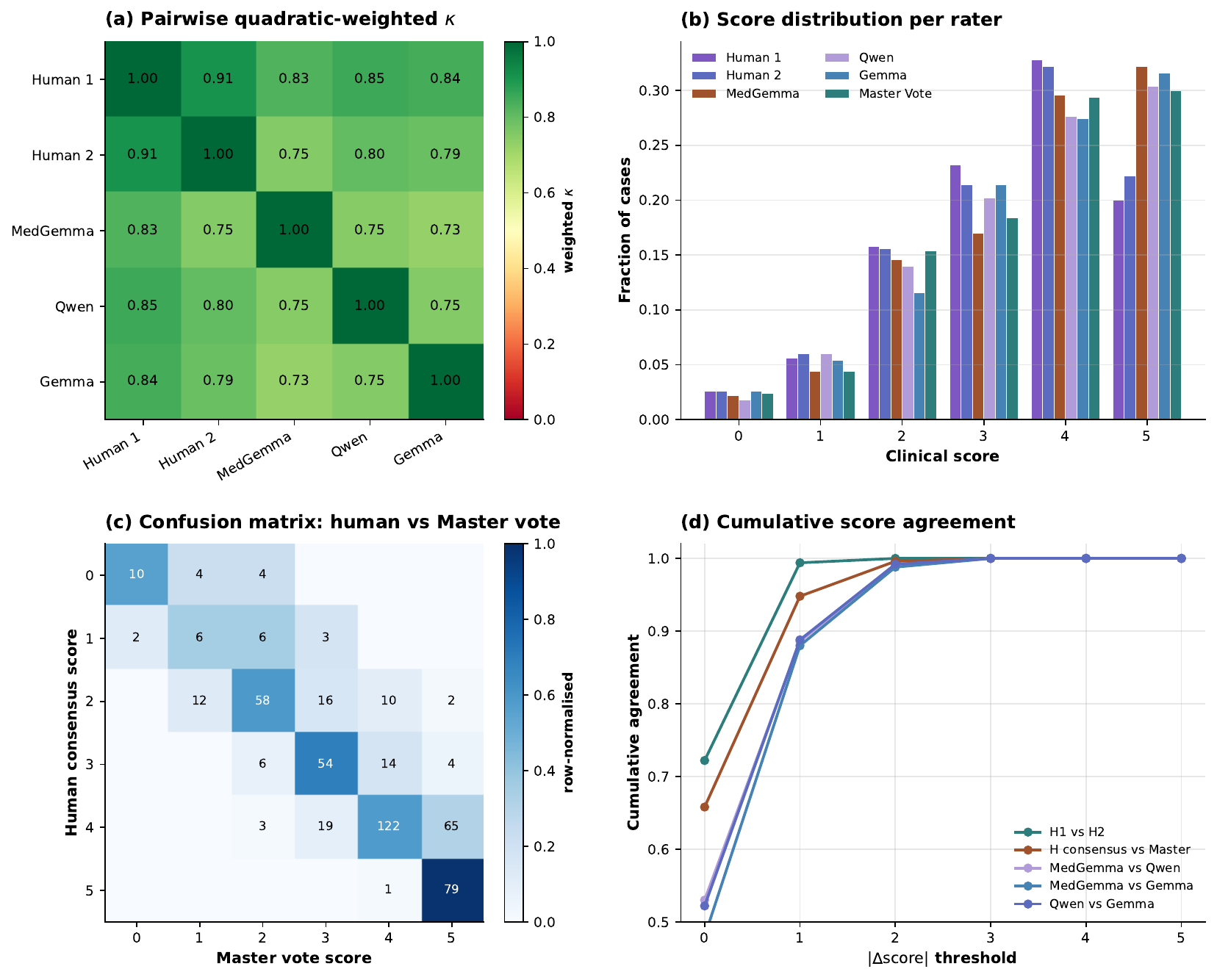}
    \caption{Inter-annotator agreement suite on the $N=500$ paired evaluation sample. (a)~Pairwise quadratic-weighted $\kappa$ across the two human annotators and the three LLM judges. The two humans agree at $\kappa{=}0.91$; every human-LLM pair lies in the substantial-to-almost-perfect band. (b)~Per-rater score distributions: humans and LLMs put most of their probability mass on the same $\{3,4,5\}$ region of the scale. (c)~Confusion matrix between the rounded human-consensus score and the LLM Master vote (row-normalised), showing concentration on and immediately adjacent to the diagonal. (d)~Cumulative agreement: for every plotted rater pair, more than $90\%$ of cases agree within $\pm 1$ score point.}
    \label{fig:appendix_iaa_grid}
\end{figure*}

\begin{table}[!t]
\centering
\small
\begin{tabular}{lcc}
\toprule
\textbf{Model} & \textbf{VinDr-CXR} & \textbf{RSNA} \\
\midrule
CheXWorld & $94.23 \pm 0.18$ & $73.21 \pm 0.34$ \\
RAD-DINO & $94.92 \pm 0.12$ & $75.64 \pm 0.63$ \\
RadJEPA & $\mathbf{95.36 \pm 0.15}$ & $\mathbf{79.12 \pm 0.21}$ \\
\bottomrule
\end{tabular}
\caption{AUROC comparison reproduced over 5 different random seeds for completeness.}
\label{tab:chexworld_auroc}
\end{table}

\section{Human Evaluation}
\label{app:human_eval}

\subsection{Inter-Annotator Agreement}
\label{sec:appendix_iaa}

\paragraph{Setup:}
Two clinical annotators (H1, H2) and three LLM judges (MedGemma-4B, Qwen-3-4B, Gemma-4B) independently graded a stratified sample of $N=500$ \radjepa{}-generated reports drawn equally from the IU-Xray and MIMIC-CXR test sets ($250$ reports each). Each report was rated along two axes: an overall clinical score $s\in\{0,\ldots,5\}$ ($5$ = clinically equivalent to the ground-truth report; $0$ = unsafe or dangerously contradicted) and a binary Acceptable flag set to $1$ iff $s \ge 4$. The three LLM judgements were combined into a single Master vote by per-metric majority.

\paragraph{Agreement statistics:}
We report Cohen's quadratic-weighted $\kappa$ for the ordinal score, Cohen's unweighted $\kappa$ for the binary flag, Fleiss' $\kappa$~\citep{fleiss1971measuring} and Krippendorff's ordinal $\alpha$ for multi-rater comparisons, and ICC(2,1)~\citep{shrout1979intraclass} for absolute-agreement reliability of the ordinal score. Following \citet{landis1977measurement}, values of $0.81$--$1.00$ are interpreted as almost above average, $0.61$--$0.80$ as substantial, and $0.41$--$0.60$ as moderate.

\begin{table}[!t]
    \centering
    \small
    \begin{tabular}{lcc}
        \toprule
        Comparison & Score (QWK) & Acceptable ($\kappa$) \\
        \midrule
        H1 vs H2                & 0.91 & 0.83 \\
        \midrule
        H1 vs LLM Master vote   & 0.88 & 0.86 \\
        H2 vs LLM Master vote   & 0.81 & 0.77 \\
        \midrule
        H1 vs MedGemma-4B       & 0.83 & 0.75 \\
        H1 vs Qwen-3-4B         & 0.85 & 0.79 \\
        H1 vs Gemma-4B          & 0.84 & 0.75 \\
        H2 vs MedGemma-4B       & 0.75 & 0.65 \\
        H2 vs Qwen-3-4B         & 0.80 & 0.70 \\
        H2 vs Gemma-4B          & 0.79 & 0.66 \\
        \bottomrule
    \end{tabular}
    \caption{Pairwise agreement on the 0--5 clinical score (quadratic-weighted Cohen's $\kappa$, QWK) and on the binary Acceptable flag (Cohen's $\kappa$). All human-LLM comparisons fall in the substantial-to-almost-perfect range, with the LLM Master vote tracking the human-human reference most closely.}
    \label{tab:iaa_pairwise}
\end{table}

\begin{table}[!t]
    \centering
    \small
    \begin{tabular}{lc}
        \toprule
        Multi-rater statistic & Value \\
        \midrule
        Krippendorff's $\alpha$ (ordinal, 5 raters, score) & 0.77 \\
        ICC(2,1) (5 raters, score)                         & 0.80 \\
        Fleiss' $\kappa$ (5 raters, Acceptable)            & 0.71 \\
        Within $\pm 1$ score agreement (H vs Master vote)  & 94.2\% \\
        \bottomrule
    \end{tabular}
    \caption{Multi-rater reliability across the two human annotators and the three LLM judges. All coefficients fall in the substantial range by the Landis--Koch scale~\citep{landis1977measurement}.}
    \label{tab:iaa_multirater}
\end{table}

\paragraph{Summary:}
The two human annotators show almost-perfect agreement on both the ordinal score (QWK $=\textbf{0.91}$) and the acceptability flag ($\kappa=0.83$), establishing a high-reliability reference. The LLM majority vote tracks this reference at QWK $=\textbf{0.84}$ and $\kappa=0.81$ on average across the two humans, and the multi-rater coefficients (Krippendorff's $\alpha = 0.77$, ICC $= 0.80$, Fleiss' $\kappa = 0.71$) confirm substantial agreement across the full panel. These results support the use of the majority-vote LLM protocol as a scalable and clinically meaningful complement to human evaluation. The same conclusions are visible in Figure~\ref{fig:appendix_iaa_grid}, which shows the pairwise weighted-$\kappa$ matrix,  per-rater score distributions, the human-consensus to vote confusion matrix, and cumulative agreement curves.

\paragraph{Qualitative Error Analysis:}
\label{app:qualitative_errors}

Aggregate report-generation metrics provide limited insight into the specific clinical statements responsible for individual generation errors. We therefore complement the quantitative evaluation with a case-level qualitative analysis of representative reports from the 500-report human-reviewed sample. Table~\ref{tab:qualitative_errors} organizes these observations into four clinically relevant categories: false-positive findings, hallucinated clinical content, missed subtle findings, and low-frequency disease categories. We show only the clinically relevant portions of the reference and generated reports to make each discrepancy directly interpretable. These examples are intended to characterize representative generation behaviours rather than estimate the prevalence of individual error types.

\begin{table*}[!t]
\centering
\small
\setlength{\tabcolsep}{2pt}
\renewcommand{\arraystretch}{1.5}

\begin{tabular}{
p{2.0cm}
p{4.45cm}
p{4.45cm}
p{4.55cm}
}
\toprule
\textbf{Category} &
\textbf{Ground Truth} &
\textbf{Generated Report} &
\textbf{Human Interpretation} \\
\midrule

\textbf{False positive}
&
No focal air-space opacity. No pleural effusion or pneumothorax.
&
Mild bibasilar atelectatic change with a \textbf{small left pleural effusion}.
&
A clinically plausible pleural abnormality is reported despite its absence in the reference.
\\
\cmidrule(lr){2-4}

&
Cardiomediastinal silhouette is within normal limits. No pulmonary vascular congestion or focal consolidation.
&
Mild cardiomegaly with \textbf{interstitial pulmonary edema}.
&
The generation introduces both cardiac enlargement and edema in an otherwise negative reference report.
\\
\cmidrule(lr){2-4}

&
Lungs are clear without focal consolidation. No acute cardiopulmonary abnormality.
&
There is a \textbf{small right lower-lobe opacity}, likely representing early air-space disease.
&
A focal pulmonary finding is asserted despite no corresponding abnormality in the reference.
\\
\midrule

\textbf{Hallucinated finding}
&
No focal pulmonary opacity, pleural effusion, or pneumothorax.
&
No focal consolidation. A \textbf{right internal jugular central venous catheter terminates in the lower SVC}.
&
The generation introduces a support device that is not present in the reference information.
\\
\cmidrule(lr){2-4}

&
Stable cardiomediastinal contours with mild bibasilar atelectatic change.
&
Stable postoperative appearance following \textbf{median sternotomy and CABG}, with mild bibasilar atelectasis.
&
The pulmonary finding is compatible with the reference, but unsupported surgical history is added.
\\
\cmidrule(lr){2-4}

&
Mild left basilar opacity, likely atelectatic. No pleural effusion.
&
Left basilar atelectatic opacity is present. \textbf{Postoperative changes following left upper lobectomy are again noted}.
&
A valid image finding is accompanied by unsupported procedure-related clinical context.
\\
\midrule

\textbf{Missed subtle finding}
&
Heart size is moderately enlarged. Mild left costophrenic angle blunting is present. Impression includes a \textbf{small left pleural effusion versus pleural thickening}, with streaky and patchy bibasilar opacities.
&
Moderate cardiomegaly and left costophrenic angle blunting are described, together with bibasilar streaky opacities; however, the report concludes, ``There is \textbf{no pleural effusion or pneumothorax}.''
&
Several accompanying abnormalities are retained, but the small left pleural abnormality is explicitly negated.
\\
\cmidrule(lr){2-4}

&
The heart size and pulmonary vascularity are within normal limits. A \textbf{right pleural effusion is present and appears increased}; the impression again identifies a small increased right pleural effusion.
&
The heart and mediastinum are described as unremarkable, followed by: ``There is no focal consolidation, \textbf{effusion}, or pneumothorax.''
&
The right pleural effusion is omitted and directly contradicted by an explicit negative statement.
\\
\cmidrule(lr){2-4}

&
Stable elevation of the right hemidiaphragm with questionable increased right basilar air-space opacity. The impression additionally reports a \textbf{small right pleural effusion}.
&
The generated report recovers the elevated right hemidiaphragm and right basilar air-space opacity, but states: ``No pneumothorax or \textbf{pleural effusion}.''
&
Neighbouring radiographic findings are preserved while the subtle pleural abnormality is missed.
\\
\midrule

\textbf{Less common finding}
&
Small-volume \textbf{pneumomediastinum} with mild associated subcutaneous emphysema. No pneumothorax.
&
Subcutaneous emphysema is present along the chest wall. No pneumothorax or pleural effusion.
&
A related abnormality is recognized, while the less commonly represented pneumomediastinum finding is omitted.
\\
\cmidrule(lr){2-4}

&
Acute mildly displaced \textbf{right posterior sixth-rib fracture}. No pneumothorax.
&
No focal pulmonary abnormality, pleural effusion, or pneumothorax. \textbf{No acute osseous abnormality}.
&
The focal osseous abnormality is missed and additionally contradicted by the generated report.
\\
\cmidrule(lr){2-4}

&
Extensive \textbf{subcutaneous emphysema} over the right chest wall and lower neck without pneumothorax.
&
Right chest-wall and lower-neck \textbf{subcutaneous emphysema} is present. No pneumothorax.
&
The example illustrates successful recovery of a less commonly represented finding rather than an error.
\\

\bottomrule
\end{tabular}

\caption{Examples from false-positive findings, hallucinated clinical content, missed subtle abnormalities, and behaviour on low-frequency disease categories. Reference and generated reports are shortened to the clinically relevant content, and \textbf{bold text} marks the key finding responsible for the agreement or discrepancy.}
\label{tab:qualitative_errors}
\end{table*}

\begin{figure*}[!t]
    \centering
    \includegraphics[width=\textwidth]{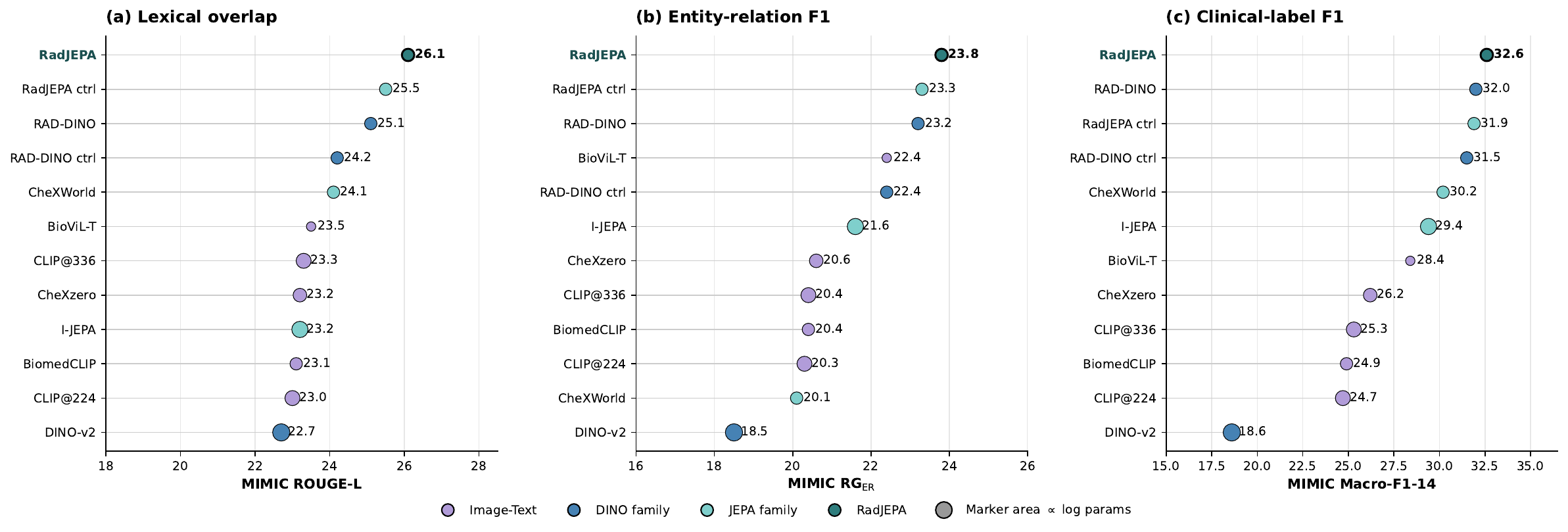}
    \caption{Per-metric ranking of frozen encoders on the MIMIC-CXR report-generation benchmark (cf.\ Table~\ref{tab:findings_generation}). Within each panel, encoders are sorted from best to worst on the corresponding metric. Marker area encodes the backbone parameter count on a log scale; colour encodes the model family. \radjepa{} sits at the top of every panel.}
    \label{fig:appendix_efficiency_pareto}
\end{figure*}

\section{GREEN Score Evaluation}
\label{sec:appendix_green}

In addition to the lexical and clinical-NLP metrics reported in the main paper, we also evaluate the generated reports with the GREEN score~\citep{ostmeier2024green}, a language-model-based clinical-correctness metric that scores a generated report against the reference using a fine-tuned LLM judge. The metric is bounded in $[0, 1]$, with higher values indicating better clinical agreement.

\begin{table}[!t]
\centering
\small
\begin{tabular}{lcc}
\toprule
\textbf{Model} & \textbf{MIMIC-CXR} & \textbf{IU-Xray} \\
\midrule
CheXWorld  & 0.471 & 0.498 \\
\raddino{}             & 0.479 & 0.508 \\
I-JEPA             & 0.455 & 0.482 \\
\radjepacontrol{} & 0.457 & 0.486 \\
\radjepa{}        & \textbf{0.497} & \textbf{0.525} \\
\bottomrule
\end{tabular}
\caption{GREEN score~\citep{ostmeier2024green} comparison across image-only encoders on MIMIC-CXR and IU-Xray. All numbers are computed under our evaluation pipeline using the publicly released GREEN model. Higher is better; bold marks the best value per column.}
\label{tab:green_scores}
\end{table}

Table~\ref{tab:green_scores} shows that \radjepa{} achieves the highest GREEN score on both datasets, with absolute margins of $\textbf{+0.018}$ on MIMIC-CXR and $\textbf{+0.017}$ on IU-Xray over \raddino{}, the next-best baseline. The relative ranking of the remaining encoders is consistent with the ranking observed under the lexical and clinical-NLP metrics in Table~\ref{tab:findings_generation} and under the inter-annotator analysis in Appendix~\ref{sec:appendix_iaa}, which provides a third, independent line of evidence that the gains reported in the main paper reflect genuine improvements in clinical report quality rather than artefacts of any single metric family.

\section{Per-Metric Model Ranking}
\label{sec:appendix_pareto}

Figure~\ref{fig:appendix_efficiency_pareto} ranks every encoder evaluated in Table~\ref{tab:findings_generation} on the three MIMIC report-quality metrics. Models are sorted within each panel from best to worst on the corresponding metric; the marker area encodes the backbone parameter count on a log scale, so the visualisation also exposes the efficiency profile of each encoder. \radjepa{} sits at the top of every panel despite a $86$M backbone and a $224 \times 224$ input resolution, while the next-best models, \raddinoControl{} and \raddino{}, require approximately $5.4 \times$ more input pixels per image. DINO-v2 occupies the last row of every panel even though it carries $1.1$B parameters, confirming that the gain reported in the main table is not explained by either capacity alone.

\section{Visual Analysis}
\label{sec:appendix_analysis}

\subsection{Report Generation Consistency}
\label{sec:appendix_report_generation}

\begin{figure*}[!t]
    \centering
    \includegraphics[width=\textwidth]{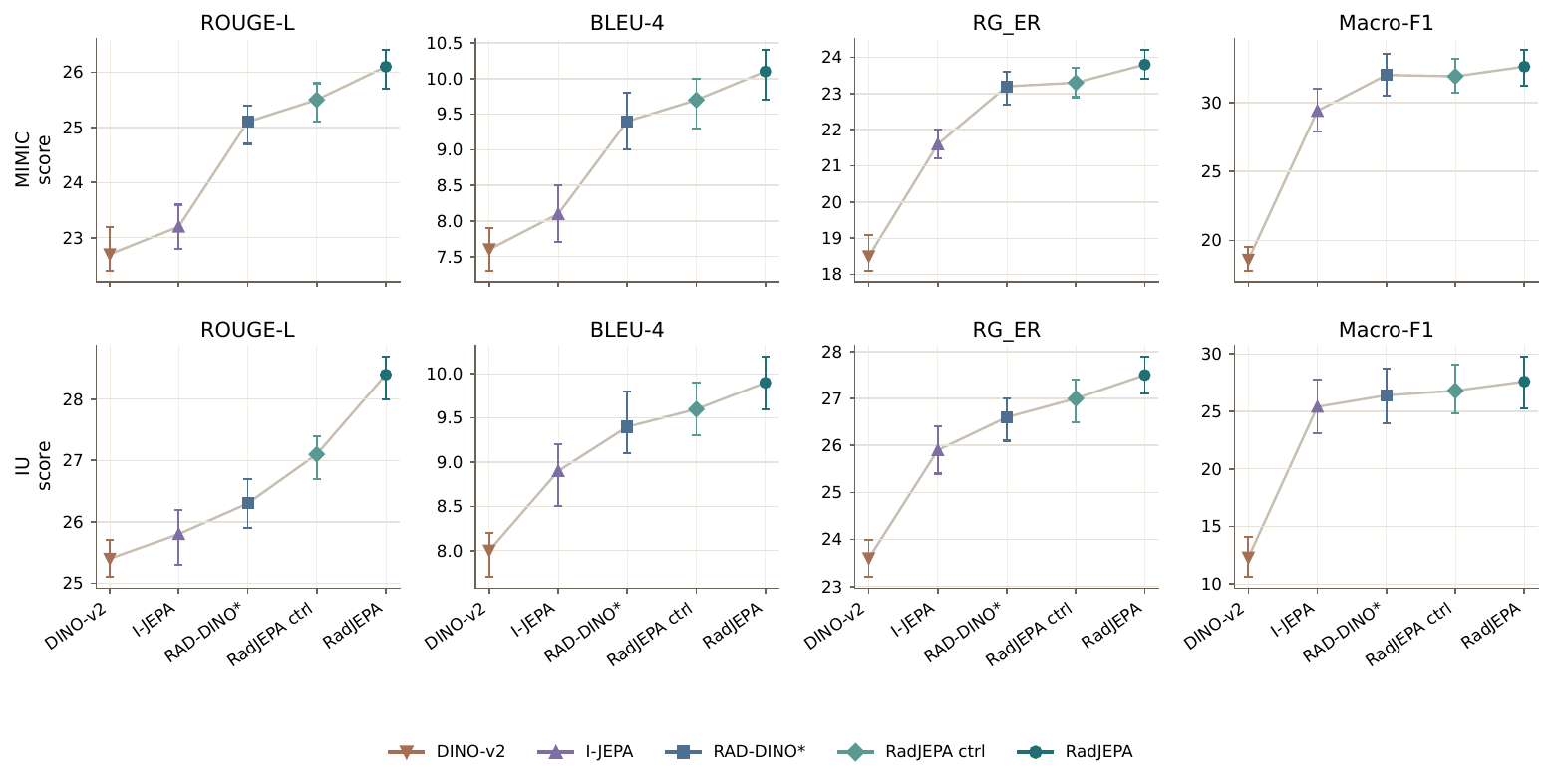}
    \caption{
    Report generation medians with 95\% confidence intervals. Each row is one dataset and each column is one metric.
    }
    \label{fig:appendix_report_generation_grid}
\end{figure*}

The report generation results show a consistent pattern across both datasets. RadJEPA is strongest on every plotted metric, and the gains are not explained by token count alone, since DINO-v2, I-JEPA, \raddino{}, RadJEPA\textsubscript{control}, and RadJEPA all use 1369 visual tokens. Narrative metrics and clinical F1 improve together, suggesting that the encoder improves not only lexical overlap but also disease-level content. Confidence intervals are narrow for ROUGE-L, BLEU-4, and RG\textsubscript{ER}, while Macro-F1 is wider, especially on IU-Xray. (cf.\ Figure \ref{fig:appendix_report_generation_grid}, Table \ref{tab:appendix_report_generation_summary})

\begin{table}[!t]
    \centering
    \small
    \begin{adjustbox}{width=0.98\columnwidth}
    \begin{tabular}{lccccc}
        \toprule
        Model & Input & Tokens & MIMIC & IU & Mean CI width \\
        \midrule
        DINO-v2 & 518$\times$518 & 1369 & 16.85 & 17.32 & 1.19 \\
        I-JEPA & 224$\times$224 & 1369 & 20.57 & 21.50 & 1.60 \\
        \raddino{} & 518$\times$518 & 1369 & 22.43 & 22.18 & 1.56 \\
        RadJEPA\textsubscript{control} & 224$\times$224 & 1369 & 22.60 & 22.62 & 1.40 \\
        RadJEPA & 224$\times$224 & 1369 & \textbf{23.15} & \textbf{23.35} & 1.42 \\
        \bottomrule
    \end{tabular}
    \end{adjustbox}
    \caption{
    Report generation summary. Mean CI width averages the reported interval widths over both datasets and all four metrics.
    }
    \label{tab:appendix_report_generation_summary}
\end{table}

\subsection{Portability Across VLM Backbones}
\label{sec:appendix_vlm_portability}

\begin{figure*}[!t]
    \centering
    \includegraphics[width=\textwidth]{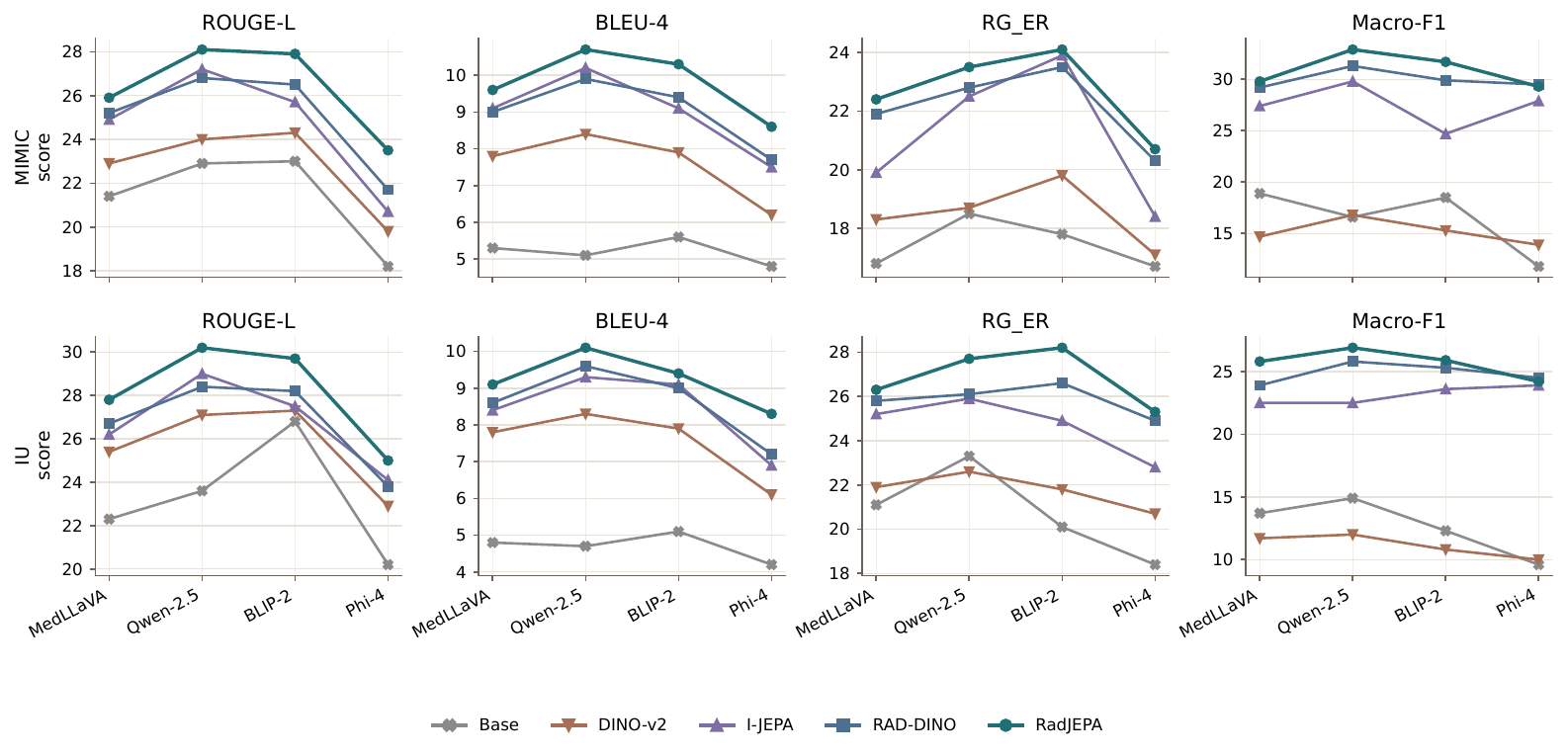}
    \caption{
    Encoder replacement results across VLM backbones. Lines track the same encoder across MedLLaVA, Qwen-2.5, BLIP-2, and Phi-4.
    }
    \label{fig:appendix_vlm_grid}
\end{figure*}

Replacing the vision encoder across four VLM backbones provides a stronger test than a single decoder setting. RadJEPA leads in nearly all cells, including the weaker Phi-4 setting. Gains are most consistent for ROUGE, BLEU, and RG\textsubscript{ER}, while Macro-F1 is more decoder-sensitive, with RAD-DINO narrowly ahead for Phi-4. Thus, the encoder consistently improves report language and entity recovery, while disease-label extraction still depends on decoder use of the visual features. (Figure \ref{fig:appendix_vlm_grid}, Table \ref{tab:appendix_vlm_summary})

\subsection{Classification Transfer Profile}
\label{sec:appendix_classification_profile}

\begin{figure*}[!t]
    \centering
    \includegraphics[width=1.0\textwidth]{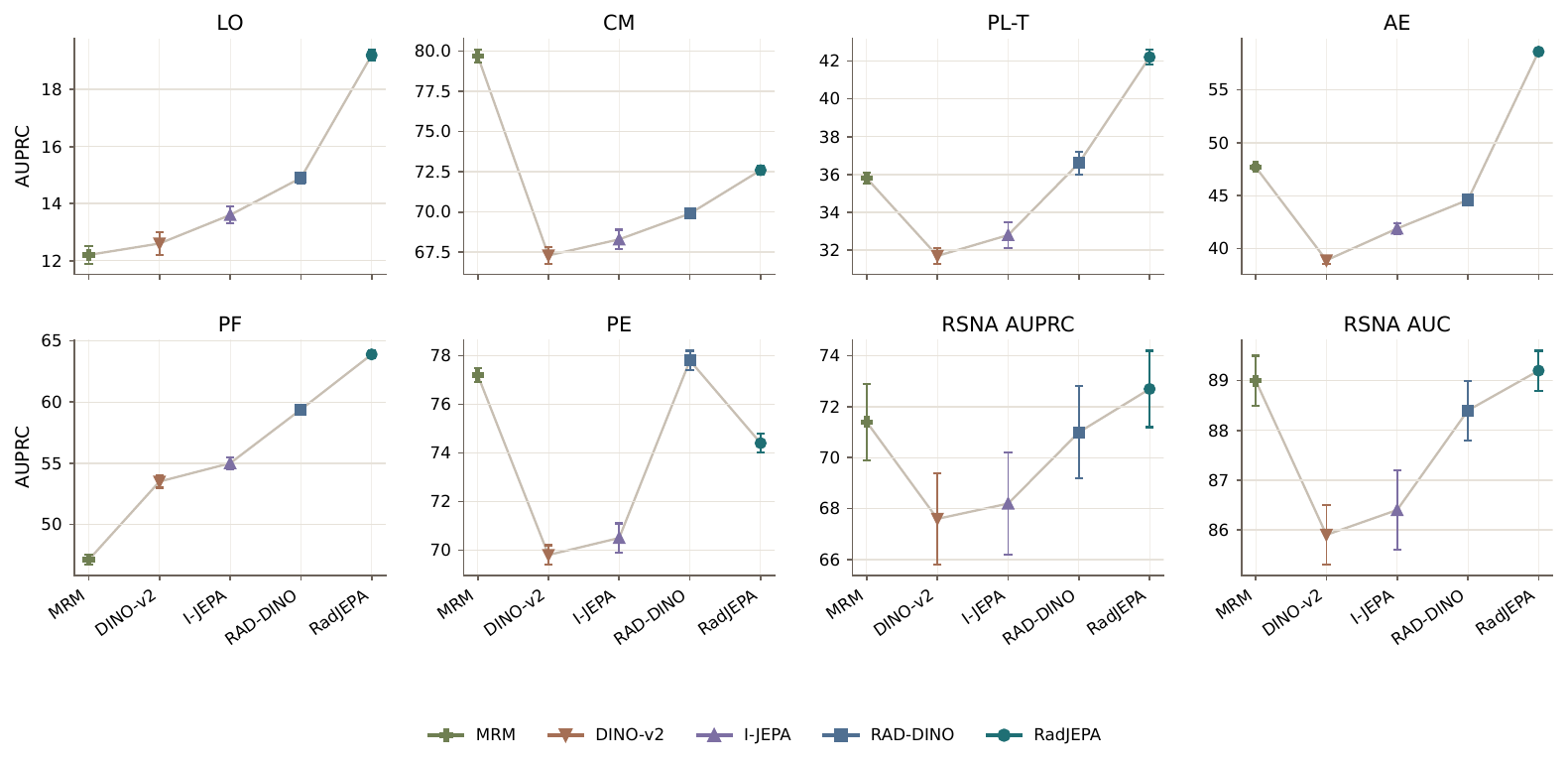}
    \caption{
    Classification performance with reported standard deviations. The six VinDr-CXR panels use AUPRC, followed by RSNA AUPRC and RSNA AUC.
    }
    \label{fig:appendix_classification_grid}
\end{figure*}

The classification profile is label-dependent rather than uniformly superior across diseases. RadJEPA is strongest for lung opacity, pleural thickening, aortic enlargement, and pulmonary fibrosis, while MRM remains strongest for cardiomegaly and RAD-DINO for pleural effusion. This highlights the distinction between broad transfer strength and label-specific performance. RadJEPA also shows low run-to-run variation on VinDr-CXR, with standard deviations of 0.2–0.4 across the six displayed classes. (cf.\ Figure \ref{fig:appendix_classification_grid}, Table \ref{tab:appendix_classification_profile})

\begin{table*}[!t]
    \centering
    \small
    \renewcommand{\arraystretch}{0.75} 
    \setlength{\tabcolsep}{6pt}        

    \begin{tabular*}{\textwidth}{@{\extracolsep{\fill}}llrrr@{}}
        \toprule
        Target & Strongest non-RadJEPA & RadJEPA & Comparator & Margin \\
        \midrule
        Lung opacity & RAD-DINO & 19.2 & 14.9 & +4.3 \\
        Cardiomegaly & MRM & 72.6 & 79.7 & -7.1 \\
        Pleural thickening & RAD-DINO & 42.2 & 36.6 & +5.6 \\
        Aortic enlargement & \raddino{} & 58.6 & 48.5 & +10.1 \\
        Pulmonary fibrosis & RAD-DINO & 63.9 & 59.4 & +4.5 \\
        Pleural effusion & RAD-DINO & 74.4 & 77.8 & -3.4 \\
        RSNA AUPRC & MRM & 72.7 & 71.4 & +1.3 \\
        RSNA AUC & MRM & 89.2 & 89.0 & +0.2 \\
        \bottomrule
    \end{tabular*}

    \caption{
    Target-wise comparison against the strongest non-RadJEPA model for each endpoint.
    }
    \label{tab:appendix_classification_profile}
\end{table*}

\begin{table}[H]
    \centering
    \small
    \begin{adjustbox}{width=0.75\columnwidth}
    \begin{tabular}{lc}
        \toprule
        Metric & Mean RadJEPA $-$ RAD-DINO \\
        \midrule
        ROUGE-L & +1.35 \\
        BLEU-4 & +0.71 \\
        RG\textsubscript{ER} & +0.79 \\
        Macro-F1-14 & +0.89 \\
        \midrule
        All metrics & +0.93 \\
        \bottomrule
    \end{tabular}
    \end{adjustbox}
    \caption{
    Mean performance margins of RadJEPA over RAD-DINO across report-generation metrics, evaluated under the same VLM and dataset settings. Positive values indicate improvements with RadJEPA, with an average gain of +0.93 across all metrics.
    }
    \label{tab:appendix_vlm_summary}
\end{table}

\subsection{Segmentation Decoder Scaling}
\label{sec:appendix_segmentation_scaling}

\begin{figure*}[p]
    \centering
    \includegraphics[width=\textwidth]{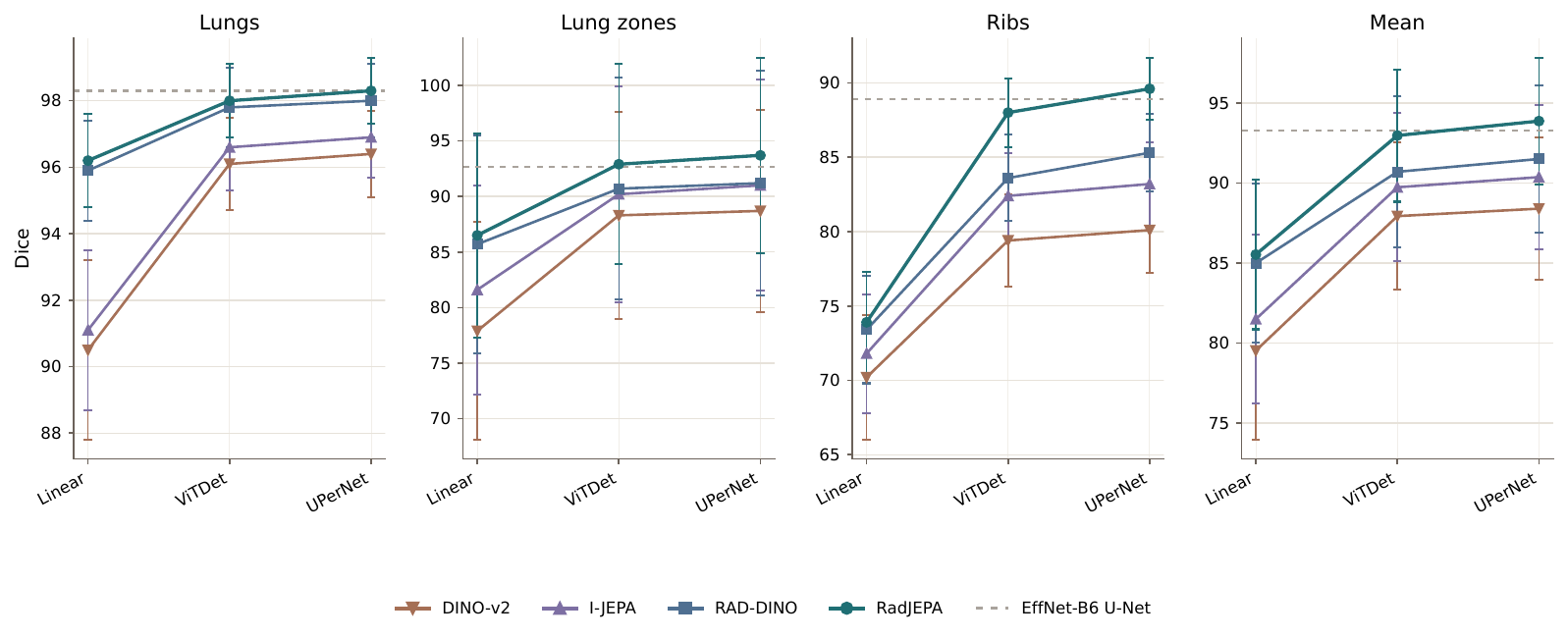}
    \caption{
    Dice scores across decoder families with reported standard deviations.
    The dashed reference line marks the EfficientNet-B6 U-Net value for each panel.
    }
    \label{fig:appendix_segmentation_grid}
\end{figure*}

The segmentation results show how the representation behaves as the decoder becomes stronger. With a linear decoder, RadJEPA is only slightly ahead of RAD-DINO on mean Dice. With ViTDet and UPerNet, the separation becomes much clearer, especially for lung zones and ribs. Lungs are already close to saturation for most strong models, so the more informative targets are the harder anatomical structures. RadJEPA-UPerNet also has lower reported standard deviations than RAD-DINO-UPerNet on lungs, lung zones, and ribs, which makes the gain look stable rather than noisy. (cf.\ Figure \ref{fig:appendix_segmentation_grid}, Table \ref{tab:appendix_segmentation_summary}, \ref{tab:appendix_segmentation_anatomy_summary})

\begin{table}[H]
    \centering
    \small
    \setlength{\tabcolsep}{3.5pt}
    \begin{adjustbox}{max width=\columnwidth}
    \begin{tabular}{lrrrrr}
        \toprule
        Model & Lin. & ViTDet & UPer. & SD & $\Delta$ \\
        \midrule
        DINO-v2 & 79.5 & 87.9 & 88.4 & 4.4 & +8.9 \\
        I-JEPA & 81.5 & 89.7 & 90.4 & 4.5 & +8.9 \\
        RAD-DINO & 85.0 & 90.7 & 91.5 & 4.6 & +6.5 \\
        RadJEPA & \textbf{85.5} & \textbf{93.0} & \textbf{93.9} & \textbf{4.0} & +8.3 \\
        \bottomrule
    \end{tabular}
    \end{adjustbox}
    \caption{
    Frozen-encoder segmentation summary using mean Dice across lungs, lung zones, and ribs denotes the linear segmentation setting, ViTDet denotes the ViTDet decoder setting, and UPer. denotes the UPerNet decoder setting. 
    SD reports the standard deviation associated with the UPerNet Dice scores, and $\Delta$ reports the absolute gain from the linear setting to the UPerNet setting. 
    All values are reported on the 0--100 Dice scale, with higher mean Dice indicating better segmentation performance and lower SD indicating more consistent performance across the evaluated anatomical structures.
    }
    \label{tab:appendix_segmentation_summary}
\end{table}

\begin{table*}[!t]
    \centering
    \small
    \begin{adjustbox}{width=\textwidth}
    \begin{tabular}{lrrrrr}
        \toprule
        Anatomy & RadJEPA-UPerNet & RAD-DINO-UPerNet & EfficientNet-B6 U-Net & RadJEPA $-$ RAD-DINO & RadJEPA SD \\
        \midrule
        Lungs & 98.3 & 98.0 & 98.3 & +0.3 & 1.0 \\
        Lung zones & 93.7 & 91.2 & 92.7 & +2.5 & 8.8 \\
        Ribs & 89.6 & 85.3 & 88.9 & +4.3 & 2.1 \\
        \bottomrule
    \end{tabular}
    \end{adjustbox}
    \caption{
    Anatomy-level UPerNet comparison. The larger separation appears on lung zones and ribs, where the task is less saturated than whole-lung segmentation.
    }
    \label{tab:appendix_segmentation_anatomy_summary}
\end{table*}

\begin{table*}[!t]
\centering
\scriptsize

\setlength{\tabcolsep}{3.2pt}
\renewcommand{\arraystretch}{1.28}
\setlength{\arrayrulewidth}{0.35pt}
\setlength{\doublerulesep}{1.2pt}

\begin{adjustbox}{width=\textwidth}
\begin{tabular}{
|>{\raggedright\arraybackslash}p{2.35cm}
|>{\raggedright\arraybackslash}p{3.65cm}
|>{\raggedright\arraybackslash}p{3.55cm}
|>{\raggedright\arraybackslash}p{3.85cm}
|>{\raggedright\arraybackslash}p{3.65cm}|
}

\hline
\textbf{Training choice}
&
\textbf{\radjepa{} (ours)}
&
\textbf{I-JEPA~\cite{assran2023self}}
&
\textbf{CheXWorld~\cite{yue2025chexworld}}
&
\textbf{\raddino{}~\cite{perez2025exploring}}
\\
\hline\hline

\textbf{Pretraining data}
&
$839{,}364$ open-source CXRs from BRAX, CheXpert, MIMIC-CXR,
ChestX-ray14, and PadChest.
&
ImageNet-1K ($\sim1.28$M natural images).
&
$\sim0.5$M frontal CXRs from MIMIC-CXR, CheXpert, and NIH ChestX-ray14.
&
$882{,}775$ CXRs in the public release; the paper checkpoint used a
different corpus/configuration.
\\
\hline

\textbf{Views used}
&
$635{,}086$ frontal and $204{,}278$ lateral images.
&
Natural images; one crop per sample.
&
Frontal images only.
&
Frontal and lateral CXRs in the public release.
\\
\hline

\textbf{Initialization}
&
Trained from scratch.
&
Trained from scratch.
&
Trained from scratch.
&
Continued from pretrained DINOv2 weights.
\\
\hline\hline

\textbf{Image preprocessing}
&
DICOM
&
RGB images; tensor conversion and ImageNet mean/std normalization.
&
X-ray loading and normalization implemented in the released data pipeline.
&
DICOM
\\
\hline

\textbf{Image augmentation}
&
Random resized crop, scale $0.3$--$1.0$; no flip, color distortion, or
Gaussian blur.
&
Same released transform.
&
Random-resized crops plus brightness, contrast, gamma, and Gaussian-blur
transformations for domain-transition modeling.
&
Random-resized multi-crop, horizontal flip, color jitter, grayscale, and
Gaussian blur.
\\
\hline\hline

\textbf{Optimizer}
&
AdamW, $\beta=(0.9,0.999)$; no decay on bias/1-D parameters.
&
AdamW, $\beta=(0.9,0.999)$; no decay on bias/1-D parameters.
&
AdamW, $\beta=(0.9,0.999)$.
&
AdamW, $\beta=(0.9,0.999)$, with layer-wise decay and patch-embedding LR
multiplier.
\\
\hline

\textbf{Learning-rate schedule}
&
$2\times10^{-4}\rightarrow10^{-3}$ over $40$ warmup epochs, then cosine
to $10^{-6}$.
&
Same released ViT-H/14 config.
&
Linear warmup for $40$ epochs to $2\times10^{-4}$, then cosine to
$10^{-6}$.
&
Base LR $10^{-3}$ in the CXR override, $5$ warmup epochs, then cosine
to $10^{-6}$.
\\
\hline

\textbf{Weight decay}
&
Cosine $0.04\rightarrow0.4$.
&
Cosine $0.04\rightarrow0.4$.
&
Fixed at $0.05$ in the released command.
&
Cosine $0.04\rightarrow0.4$.
\\
\hline

\textbf{EMA / teacher schedule}
&
Linear $0.996\rightarrow1.0$.
&
Linear $0.996\rightarrow1.0$.
&
Cosine $0.996\rightarrow1.0$.
&
Teacher momentum $0.992\rightarrow1.0$.
\\
\hline\hline

\textbf{Epochs / updates}
&
$300$ epochs.
&
$300$ epochs.
&
$300$ epochs.
&
$100$-epoch/iteration-based released configuration; public checkpoint
selected at $35$K iterations from a $100$K-iteration run.
\\
\hline

\textbf{Numerical precision}
&
bfloat16 autocast.
&
bfloat16 autocast.
&
bfloat16 autocast with automatic mixed precision.
&
fp16 mixed precision with FSDP.
\\
\hline

\textbf{Gradient clipping}
&
None.
&
None in released code.
&
Global norm clipped at $1.0$.
&
Global norm clipped at $3.0$.
\\
\hline

\textbf{Random seed}
&
$0$
&
$0$ in released training code.
&
$0$ in released command.
&
$0$ in released configuration.
\\
\hline

\end{tabular}
\end{adjustbox}

\caption{
Comparison of pretraining configurations for \radjepa{} and related
self-supervised representation-learning methods. We compare training data,
input views, initialization, augmentation, optimization, scheduling, and
numerical settings.
}
\label{tab:pretraining_optimization_comparison}
\end{table*}

\clearpage

\makeatletter
\setlength{\@dblfptop}{0pt}
\setlength{\@dblfpbot}{0pt plus 1fil}
\makeatother

\clearpage

\end{document}